\documentclass{article}
\usepackage[preprint]{neurips_2026}
\usepackage[utf8]{inputenc}
\usepackage[T1]{fontenc}
\usepackage{hyperref}
\usepackage{url}
\usepackage{booktabs}
\usepackage{multirow}
\usepackage{amsmath,amssymb}
\usepackage{amsthm}
\usepackage{graphicx}
\usepackage{xcolor}
\usepackage{array}
\usepackage{colortbl}
\usepackage{float}
\usepackage{subcaption}
\usepackage{wrapfig}
\usepackage{placeins}
\usepackage[font=small,labelfont=bf]{caption}
\title{Spatial Blindness in Whole-Slide Multiple Instance Learning}

\author{
Xiangyu Li\\
College of Intelligence and Computing\\
Tianjin University\\
Tianjin, China\\
\texttt{xiangyuli@tju.edu.cn}
\And
Ran Su\thanks{Corresponding author.}\\
College of Intelligence and Computing\\
Tianjin University\\
Tianjin, China\\
\texttt{ran.su@tju.edu.cn}
}

\newcommand{\sg}{\text{sg}}

\newtheorem{proposition}{Proposition}

\begin{document}
\raggedbottom

\maketitle

\begin{abstract}
Whole-slide MIL models are often called context-aware once graphs, Transformers, or state-space modules are placed above patch embeddings. We show that this label can be deceptive. On pathology tasks where tissue architecture is part of the diagnostic signal, several strong MIL baselines retain nearly unchanged slide-level AUC after patch coordinates are permuted. Their predictions are accurate, but largely compositional. We refer to this failure mode as \textit{spatial blindness}. Our explanation is optimization-based: dense appearance statistics are learned early under slide-level supervision, leaving weak gradients for sparse spatial relations. ResTopoMIL addresses the issue by first fitting a permutation-invariant prototype histogram and then freezing it while a lightweight graph branch learns the residual under a coordinate-shuffling constraint. The architecture is simple by design; the intervention is in how the spatial branch is trained. Across 9 public WSI benchmarks, ResTopoMIL improves classification and survival prediction with 1.15M parameters, restores sensitivity to coordinate perturbation, and gives stronger localization evidence on CAMELYON-16.
\end{abstract}

\section{Introduction}

Computational pathology increasingly learns diagnostic and prognostic models directly from digitized whole-slide images (WSIs), where cellular morphology, tissue architecture, and long-range context are preserved at gigapixel scale \citep{pantanowitz2011review,verghese2023computational,song2023artificial}. The weak supervision problem is severe: a slide may contain tens of thousands of relevant and irrelevant regions, but most clinical datasets provide only a slide-level label. Multiple instance learning (MIL) is therefore the standard formulation for WSI analysis \citep{dietterich1997solving,maron1997framework,campanella2019clinical,lu2021data}: patches are treated as instances, the slide as a bag, and the model predicts the bag label without patch-level annotation.

MIL works remarkably well in pathology, but its success can hide a limitation. Many labels are not determined by local appearance alone. Gleason grading depends on gland formation; breast cancer subtyping distinguishes ductal growth from single-file strands; prognosis often reflects invasive fronts, immune aggregates, papillary cores, solid growth, or tumor-stroma organization \citep{quail2013microenvironmental,cheng2021computational,wang2021predicting}. These are spatial statements. They depend not just on which tissue components appear, but on how they are arranged.

\begin{figure}[H]
    \centering
    \includegraphics[width=0.94\textwidth]{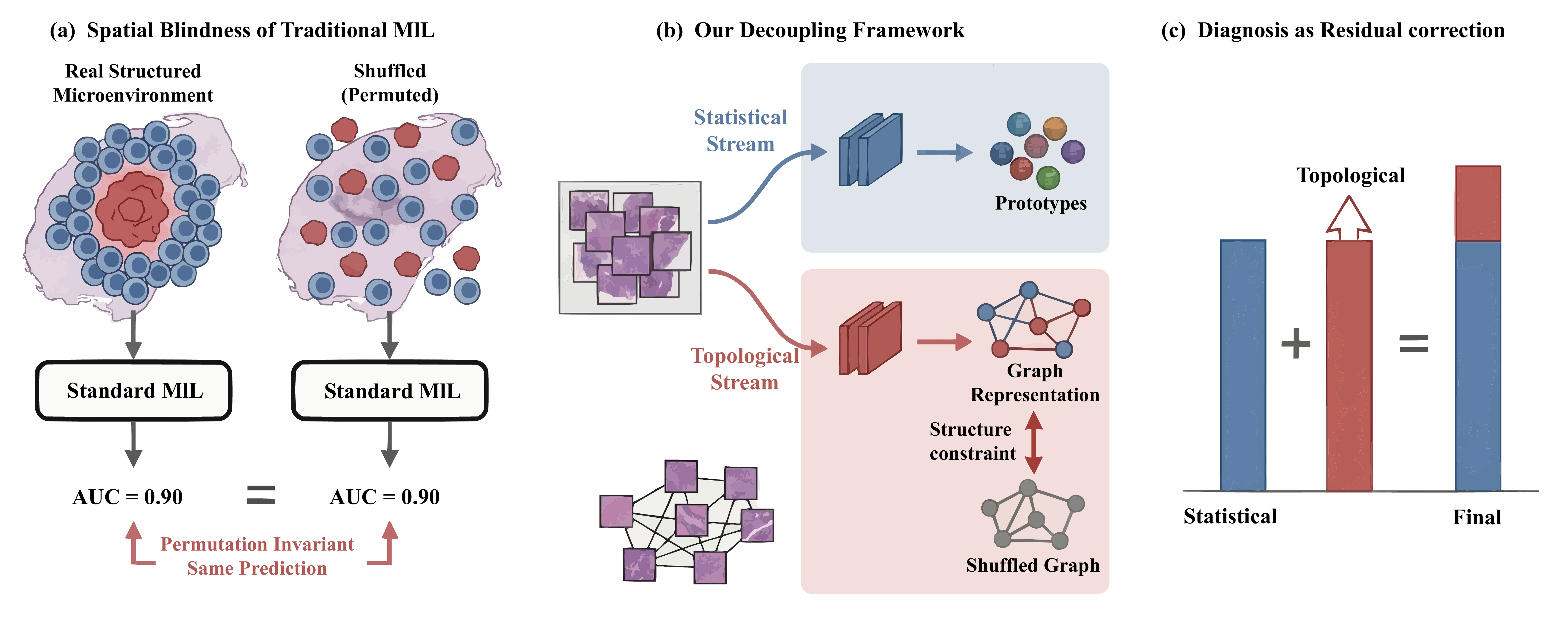}
    \caption{\textbf{The ResTopoMIL Concept.}
    (a) A standard MIL model may give similar predictions before and after spatial permutation, indicating that it mainly uses composition.
    (b) ResTopoMIL separates the problem into a statistical stream and a topological stream.
    (c) The statistical stream provides a base prediction, while the topological stream learns a residual correction from spatial organization.}
    \label{fig:teaser}
\end{figure}

At first glance, recent context-aware MIL methods should address this issue. Graph networks, Transformers, hierarchical models, and state-space models all process a slide as more than an unordered bag \citep{chen2021whole,pati2022hierarchical,adnan2020representation,vaswani2017attention,shao2021transmil,chen2022scaling,guefficiently,yang2024mambamil,zhang20252dmamba}. Architecture alone, however, does not tell us what the trained predictor uses. We use a simple stress test: keep every patch embedding fixed, and randomly permute the coordinates used to build spatial context. On tasks where architecture is label-relevant, a topology-using model should suffer. Several strong context-aware baselines barely move. They have spatial machinery, but their learned decision rules behave much like bag-of-visual-words classifiers.

The question is then not whether a model contains a spatial operator, but why such an operator can remain unused. Our explanation is optimization-based. Tissue composition provides a dense, early signal: many patches contribute to the same slide label. Topological evidence is sparser and harder to align with slide-level supervision. Under joint training, the network can reduce the loss by fitting composition first; once that happens, little useful gradient remains for the spatial branch. We call this behavior \textit{optimization laziness}, in the descriptive sense that the easiest explanatory signal dominates training while the harder structural signal is left undertrained. The phenomenon is related to simplicity bias, texture bias, and gradient starvation \citep{shah2020pitfalls,geirhos2018imagenet,noroozi2016unsupervised,pezeshki2021gradient}.

ResTopoMIL follows this diagnosis. Rather than asking one network to discover composition and topology at the same time, it learns the compositional explanation explicitly. A permutation-invariant statistical stream is trained first and frozen. A lightweight graph stream is then trained on the residual, with a shuffle-based loss that asks it to distinguish real tissue topology from coordinate-permuted topology. The graph module is intentionally small. The point is not to add a heavier context block, but to change the training problem faced by the spatial branch.

This shifts the evaluation away from architecture labels. ``Context-aware'' should mean that a model's decision changes when clinically relevant spatial organization is removed. We therefore test topology-destroying coordinate perturbations, separate pure composition from pure topology in a controlled benchmark, and ask whether residual training changes both accuracy and spatial behavior. Topology-preserving transformations are discussed only as expected graph invariances, not as an additional measured benchmark.

The paper makes four contributions:
\begin{itemize}
    \item \textit{Spatial blindness} is defined and tested as insensitivity to coordinate perturbations on structure-dependent MIL tasks.
    \item A controlled composition--topology diagnostic benchmark shows that strong MIL models can solve compositional tasks while failing on pure topology.
    \item ResTopoMIL learns composition first and topology as a residual correction, with design analysis deferred to the appendix.
    \item Experiments on 9 public pathology benchmarks show gains in prediction, spatial sensitivity, and localization quality with a compact 1.15M-parameter model.
\end{itemize}

\section{Related Work}

MIL has a long history as a weak-supervision framework for learning from bag labels without instance labels \citep{dietterich1997solving,maron1997framework}. In pathology, this formulation is natural because WSIs are too large for end-to-end pixel-level training and dense annotation is expensive. Early WSI approaches used patch classifiers, clustering, global pooling, or multi-view CNN aggregation to move from local tiles to slide-level labels \citep{xu2014weakly,kraus2016classifying,das2017classifying,das2018multiple,wang2019weakly,chen2021annotation}. Attention-based MIL then became a central baseline because it learns instance weights while remaining permutation-invariant and interpretable \citep{ilse2018attention,lu2021data}. Later methods select critical instances, split bags, mine hard examples, regularize attention, or improve instance-level classifiers \citep{li2021dual,zhang2022dtfd,tang2023multiple,lin2023interventional,zhang2024attention,zhang2025aem,qu2024rethinking,shao2025multiple,zhu2025effective,zhu2023pdl}.

A second line goes beyond unordered aggregation by modeling context among patches. Graph models treat WSIs as point clouds or tissue graphs \citep{chen2021whole,adnan2020representation,pati2022hierarchical,pal2022bag}; Transformer and low-rank attention models capture long-range bag dependencies \citep{vaswani2017attention,shao2021transmil,xiong2021nystromformer,chen2022scaling,xiang2023exploring}; hierarchical and context-aware variants exploit WSI pyramids \citep{zhang2021joint,guo2023higt,buzzard2024paths,tran2025navigating,fourkioticamil,chen2024camil}; and state-space models provide efficient long-sequence modeling \citep{guefficiently,yang2024mambamil,zhang20252dmamba}. These methods matter because pathology is not i.i.d.; structured MIL already recognizes that bags with the same instances can have different labels when dependencies differ \citep{zhou2009multi,zhang2011multiple}. We ask a complementary question: after adding such modules, does the learned predictor rely on topology, or does it still mainly count visual words?

Foundation models have greatly improved pathology patch embeddings \citep{chen2023general,kang2023benchmarking,oquab2023dinov2,caron2021emerging,lu2023visual,kapse2025gecko}, while vision-language and prompt-based WSI methods broaden slide supervision \citep{shi2024vila,han2025mscpt,wongfew,tomar2025slide,gou2025queryable}. These advances make the question more urgent: strong features can make compositional shortcuts easier to exploit. ResTopoMIL therefore fixes the UNI encoder for all methods and studies the slide-level aggregation problem.

Finally, the diagnosis is related to shortcut learning and optimization bias: neural networks often prefer simple, high-variance, or dense predictive features even when structured features are available \citep{shah2020pitfalls,geirhos2018imagenet,pezeshki2021gradient}. ResTopoMIL makes the compositional signal explicit through a prototype/statistical stream related to Deep Sets, histograms, and morphological prototypes \citep{zaheer2017deep,peeples2021histogram,song2024morphological,wei2016scalable}, then trains topology as residual information. The information-theoretic connection \citep{shannon1948mathematical,cover1999elements} is used as design analysis in Section \ref{sec:theoretical_insights} and Appendix \ref{sec:theory}.

\section{Preliminaries and Motivation}
\label{sec:prelim_motivation}

A WSI is written as a bag
\begin{equation}
    X=\{(\mathbf{h}_i,\mathbf{p}_i)\}_{i=1}^{N},
\end{equation}
where $\mathbf{h}_i\in\mathbb{R}^{d}$ is a patch embedding and $\mathbf{p}_i\in\mathbb{R}^{2}$ is its slide coordinate. We distinguish two sources of label information. \textit{Composition} is the empirical distribution of patch appearances, for example the abundance of tumor-like or stromal patches. \textit{Topology} is the spatial relation among patches: clustering, gland formation, boundaries, invasive fronts, and similar architectural cues.

The distinction matters because the two signals optimize differently. Composition is dense: every patch contributes to a histogram-like summary, and slide labels are often partly predictable from the prevalence of visual phenotypes. Topology is sparse: a small set of spatial relations may carry the decisive evidence, and supervision arrives only after whole-slide aggregation. A jointly trained model is naturally drawn to the dense signal first. Once the loss has fallen, too little error may remain to train the spatial pathway. Standard validation can then look reassuring even when the explanation ignores architecture.

\pagebreak[4]
\noindent
\begin{minipage}[t]{0.54\textwidth}
\vspace{0pt}
To test whether a model uses topology, a coordinate-shuffling operator $\pi$ keeps $\{\mathbf{h}_i\}$ fixed and permutes $\{\mathbf{p}_i\}$. This preserves composition but destroys adjacency and tissue architecture. A model is defined as \textit{spatially blind} on a structure-dependent task if its prediction or performance is nearly invariant under this perturbation. Robustness is desirable when a perturbation preserves semantics; here the perturbation removes part of the diagnostic evidence.

Figure \ref{fig:evidence} shows the motivating observation. TransMIL has contextual machinery, and DS-MIL is a strong dual-stream MIL baseline; both show little AUC change after coordinate shuffling. This is not a failure of prediction. It is evidence that high slide-level AUC can be obtained from composition alone. The controlled benchmark in Section \ref{sec:synthetic_main} makes the separation explicit: strong MIL models solve a pure-composition task but fail when the label is defined only by a spatial motif.
\end{minipage}
\hfill
\begin{minipage}[t]{0.42\textwidth}
\vspace{0pt}
    \centering
    \includegraphics[width=\linewidth]{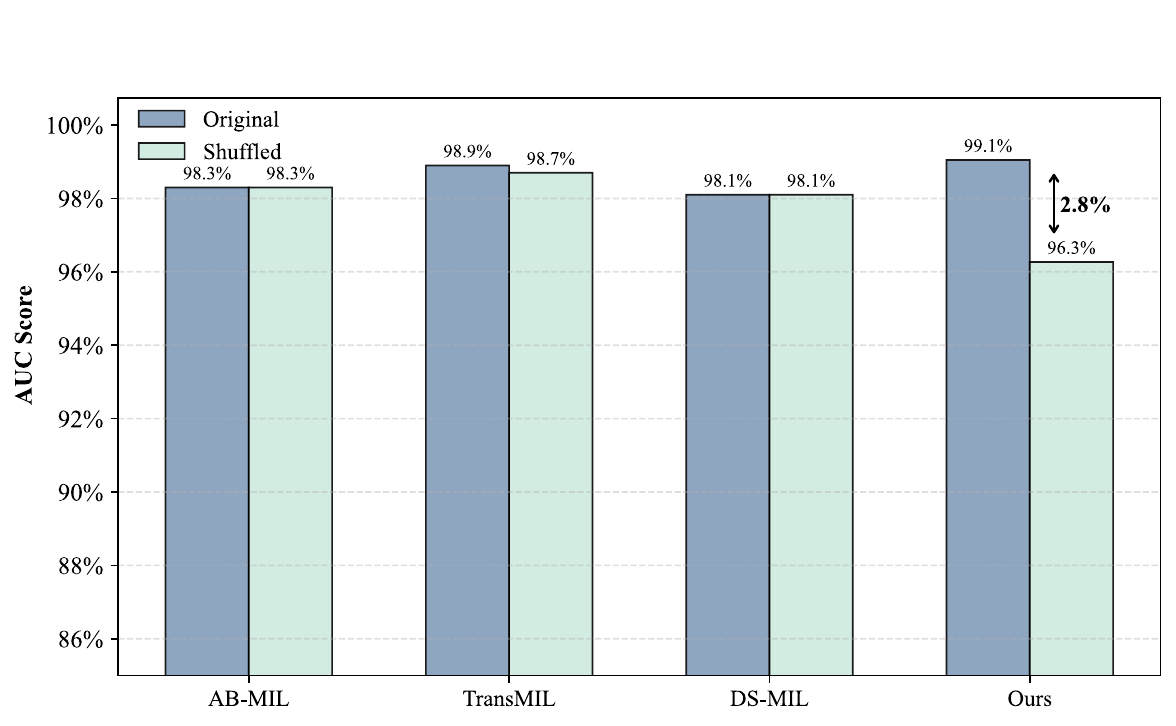}
    \captionof{figure}{\textbf{A coordinate-shuffling stress test.}
    Patch embeddings are fixed while coordinates are permuted. On TCGA-BRCA, several MIL models remain almost unchanged after complete spatial shuffling. ResTopoMIL is more sensitive to this topology-destroying perturbation, as expected when structure is label-relevant.}
    \label{fig:evidence}
\end{minipage}
\vspace{3pt}

The stress test does not imply that every pathology task must depend on topology. Some tasks are primarily compositional, and a permutation-invariant model may be the right tool. Our claim is narrower: when the label is known to depend on architecture, a model should not be invariant to a perturbation that destroys architecture while preserving patch appearances. This is the sense in which we use \textit{spatial blindness}.

This motivates the additive view
\begin{equation}
    F(X) \approx F_{stat}(\{\mathbf{h}_i\}) + F_{topo}(\{(\mathbf{h}_i,\mathbf{p}_i)\}),
    \label{eq:additive_decomp}
\end{equation}
where $F_{stat}$ captures permutation-invariant composition and $F_{topo}$ captures the remaining structure-dependent signal. The difficulty is not only architectural. If both terms are learned jointly, the first term is often easier to optimize and can reduce the loss before the second term learns. ResTopoMIL therefore makes Eq. \eqref{eq:additive_decomp} operational through staged residual training.

\section{ResTopoMIL}
\label{sec:method}

\begin{figure}[H]
    \centering
    \includegraphics[width=\linewidth]{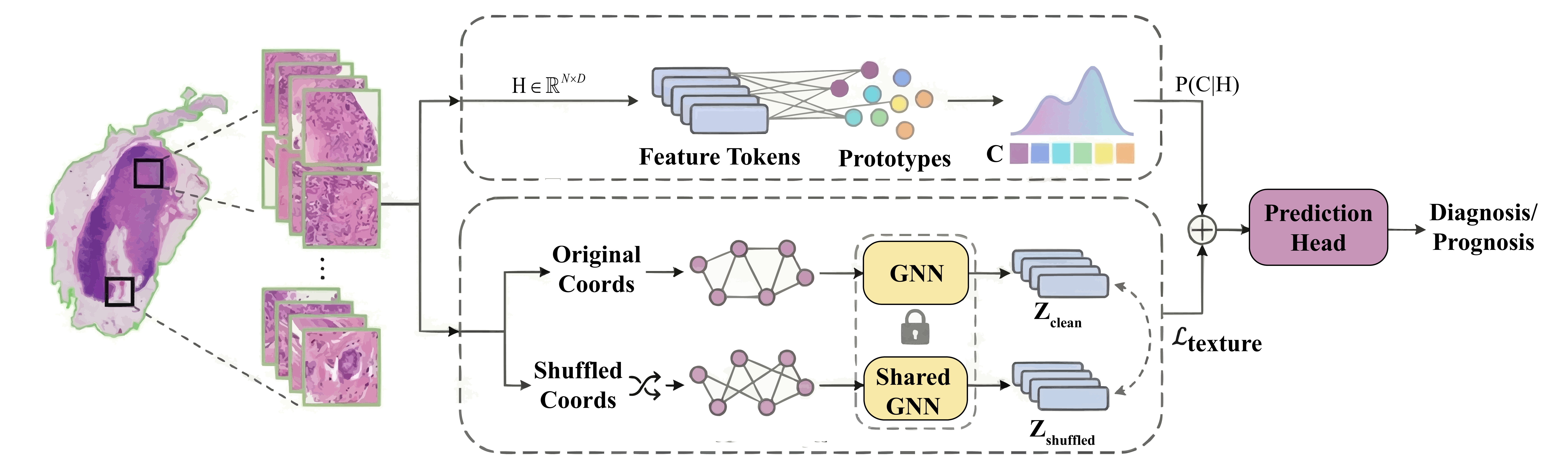}
    \caption{\textbf{Overview of the ResTopoMIL Framework.}
    The architecture decouples WSI analysis into two parallel streams.
    \textbf{Top:} The Statistical Stream captures tissue composition via a learnable prototype-based soft histogram, providing a statistical baseline.
    \textbf{Bottom:} The Topological Stream models spatial structure using a simple GNN.
    To prevent degeneration, ResTopoMIL introduces a Structure-Aware Texture Loss ($\mathcal{L}_{texture}$) that forces the model to distinguish between genuine tissue topology ($Z_{clean}$) and spatially shuffled noise ($Z_{shuffled}$).
    The final prediction is the residual summation of both streams.}
    \label{fig:arch}
\end{figure}

ResTopoMIL first learns the compositional explanation and then trains topology as a residual correction (Figure \ref{fig:arch}). The design does not hide the problem behind a larger spatial module. If spatial blindness is an optimization problem, increasing graph capacity can leave the shortcut intact. We instead use a strong permutation-invariant stream as the compositional anchor and train the graph stream on the remaining error. The residual representation is then checked against a perturbation that preserves patch appearance but destroys coordinates, so the graph branch cannot quietly become another texture aggregator.

\subsection{Statistical Anchor}
\label{sec:stat_stream}

The first stream ignores coordinates by construction. Its role is to absorb the label signal explained by tissue composition, so the second stream is not rewarded for relearning the same shortcut. We use a soft prototype histogram: richer than mean pooling, but still permutation-invariant. Let $C=\{\mathbf{c}_k\}_{k=1}^{K}$ be a learnable codebook initialized by sampled MiniBatch K-Means on randomly sampled training patch embeddings, rather than by exact K-Means over all WSI patches. For each patch embedding $\mathbf{h}_i$, the assignment is
\begin{equation}
    a_{ik}=
    \frac{\exp(-\|\mathbf{h}_i-\mathbf{c}_k\|^2/\tau)}
    {\sum_{j=1}^{K}\exp(-\|\mathbf{h}_i-\mathbf{c}_j\|^2/\tau)},
    \qquad
    \mathbf{a}_i=[a_{i1},\ldots,a_{iK}]^\top,\qquad
    \mathbf{z}_{stat}=\frac{1}{N}\sum_{i=1}^{N}\mathbf{a}_i .
\end{equation}
An MLP maps $\mathbf{z}_{stat}$ to logits $f_{stat}$. Similar to codebook-based MIL encodings \citep{wei2016scalable}, this stream measures how far one can go by counting visual phenotypes alone. Soft assignment avoids the brittleness of hard clustering, while remaining more expressive than mean pooling. A good anchor is useful scientifically as well as computationally: if the residual branch improves over it, the gain is less easily dismissed as a better bag-level composition classifier.

The anchor also gives a concrete diagnostic baseline. On a task where tissue proportion is sufficient, the residual branch should have little to add. On a task where architecture matters, the remaining error after $f_{stat}$ is precisely where gland formation, invasive fronts, or tumor-stroma organization can enter. For this reason, the graph branch is not trained as a second full classifier.

\subsection{Topological Residual Branch}
\label{sec:topo_stream}

The second stream receives the same patch embeddings together with a spatial graph. A KNN graph $\mathcal{G}=(\mathcal{V},\mathcal{E})$ is built from coordinates $\mathbf{p}_i$ and processed by a two-layer GCN:
\begin{equation}
    \mathbf{H}^{(l+1)}
    =
    \sigma\!\left(
    \tilde{\mathbf{D}}^{-\frac{1}{2}}
    \tilde{\mathbf{A}}
    \tilde{\mathbf{D}}^{-\frac{1}{2}}
    \mathbf{H}^{(l)}
    \mathbf{W}^{(l)}
    \right),
\end{equation}
where $\tilde{\mathbf{A}}=\mathbf{A}+\mathbf{I}$ and $\mathbf{H}^{(0)}$ is the matrix of patch embeddings. We use global mean pooling to obtain the graph-level representation
\begin{equation}
    \mathbf{z}_{topo}=\frac{1}{N}\sum_{i=1}^{N}\mathbf{H}^{(2)}_i,
    \qquad
    f_{topo}= \mathbf{W}_{topo}\mathbf{z}_{topo}+\mathbf{b}_{topo}.
\end{equation}
The graph is built from physical coordinates rather than feature similarity: two tumor patches can look similar while belonging to different glands or invasive fronts, whereas neighboring patches define local architecture. The branch is intentionally simple. A deeper graph model would make it harder to tell whether improvement comes from topology or from capacity; the two-layer GCN leaves the optimization question visible.

\subsection{Residual Training Objective}
\label{sec:optimization}

Training both streams from scratch reintroduces the competition that caused spatial blindness. ResTopoMIL uses two stages. Stage 1 trains only the statistical stream with standard cross-entropy, producing $f_{stat}$. Stage 2 freezes this stream and optimizes the topological branch through the combined logits
\begin{equation}
    f(X)=\mathrm{sg}[f_{stat}(X)]+f_{topo}(X),
\end{equation}
where $\mathrm{sg}[\cdot]$ stops gradients. The graph branch learns corrections to a fixed compositional predictor rather than acting as an independent classifier. The stop-gradient has two effects: the statistical stream cannot absorb the errors exposed in Stage 2, and the topological stream receives a stable residual target rather than a moving joint optimum. The residual logits can be read as structured corrections--positive when topology supports the statistical prediction, negative when spatial organization contradicts it.

To ensure that this residual is genuinely spatial, ResTopoMIL adds a shuffle-based constraint. Let $\tilde{X}$ be obtained by permuting coordinates while keeping all patch embeddings fixed. This operation preserves composition and destroys topology. Let $\tilde{\mathbf{z}}_{topo}$ denote the graph-level representation computed from this shuffled-coordinate view, and let $\mathrm{sim}(\cdot,\cdot)$ be cosine similarity. The graph representation of $X$ is required to differ from that of $\tilde{X}$:
\begin{equation}
    \mathcal{L}_{texture}
    =
    \max\!\left(0,\,
    m-\left[1-\mathrm{sim}(\mathbf{z}_{topo},\tilde{\mathbf{z}}_{topo})\right]\right).
\end{equation}
This loss is not a generic contrastive regularizer. Its negative view keeps all patch appearances unchanged, all labels unchanged, and all bag-level composition unchanged; only the coordinate-induced graph is corrupted. Satisfying the margin therefore encourages the branch to encode spatial arrangement rather than another appearance summary. This is also why the loss does not require tumor masks or pathologist-drawn regions. The final Stage-2 objective is
\begin{equation}
    \mathcal{L}_{total}
    =
    \mathcal{L}_{cls}(\mathrm{sg}[f_{stat}]+f_{topo},Y)
    +
    \lambda\mathcal{L}_{texture}.
\end{equation}
Thus the statistical stream explains composition, the residual branch explains what remains, and the auxiliary loss prevents the residual branch from becoming another permutation-invariant texture encoder.

At inference, no shuffled view is constructed. The model computes the statistical logits and topological residual logits once, then sums them. The method changes the training signal, not the deployment protocol: no patch-level annotations, topology labels, or test-time augmentations are required.

\subsection{Why Decoupling Helps}
\label{sec:theoretical_insights}

The analysis in Appendix \ref{sec:theory} is not a new general theory of gradient starvation or mutual information. Its purpose is narrower: to show why the three design choices in ResTopoMIL--a statistical anchor, stop-gradient residual training, and coordinate shuffling--belong together.

\begin{proposition}[Residual-error gating of the topological update]
\label{prop:residual_gradient}
For an additive MIL logit $f=f_{stat}+f_{topo}$ trained with cross-entropy, the topological update can be written as
\begin{equation}
    \hat p_\theta(X)=P_\theta(Y=1\mid X)=\sigma(f(X)),\qquad
    \nabla_{\theta_t}\mathcal{L}
    =
    \mathbb{E}\!\left[(\hat p_\theta(X)-Y)\nabla_{\theta_t}f_{topo}(X)\right].
\end{equation}
Hence its norm is bounded by the remaining prediction error after the current model has used the statistical shortcut:
\begin{equation}
    \|\nabla_{\theta_t}\mathcal{L}\|
    \le
    \big(\mathbb{E}(\hat p_\theta(X)-Y)^2\big)^{1/2}
    \big(\mathbb{E}\|\nabla_{\theta_t}f_{topo}(X)\|_F^2\big)^{1/2}.
\end{equation}
\end{proposition}

This proposition is deliberately modest, but it is the useful part for our setting. If composition quickly reduces the residual error, the graph branch can be present yet receive little informative supervision. Freezing $f_{stat}$ changes the learning problem: the error left by the statistical anchor becomes a stable target for the topological branch instead of a moving target shared by both branches.

\begin{proposition}[Residual branch as conditional label information]
\label{prop:residual_information}
After $Z_{stat}$ and $f_{stat}$ are fixed, optimizing the Stage-2 decoder
\begin{equation}
q_{\phi}(Y=y\mid Z_{stat},Z_{topo})
=\left[\mathrm{Softmax}\!\left(f_{stat}+f_{topo,\phi}\right)\right]_y
\end{equation}
minimizes a variational upper bound on $H(Y\mid Z_{stat},Z_{topo})$ and therefore maximizes a lower bound on the additional label information carried by topology, $I(Z_{topo};Y\mid Z_{stat})$. In the binary derivation, the same statement uses the sigmoid probability $\sigma(f_{stat}+f_{topo,\phi})$ for $Y=1$.
\end{proposition}

Proposition \ref{prop:residual_information} explains why the graph branch is trained as a residual correction rather than as another full classifier. It does not by itself make the residual spatial. That role is played by $\mathcal{L}_{texture}$: the negative view preserves all patch appearances and labels but corrupts the coordinate-induced graph, so a branch that ignores topology cannot reliably satisfy the margin. The theory therefore supports a specific method design rather than serving as a standalone theoretical contribution.

\section{Experiments}

\subsection{Experimental Setup}

Evaluation covers 9 public WSI benchmarks: BRACS \citep{brancati2022bracs}, PANDA, TCGA-NSCLC, TCGA-BRCA, and five TCGA survival cohorts (KIRC, KIRP, LUAD, STAD, UCEC). Slides are processed into non-overlapping $256\times256$ patches at 20$\times$ and represented with 1024-d UNI features. All baselines use the same features, patient-level stratified splits, preprocessing, optimizer family, and evaluation protocol; no patient appears in more than one of train, validation, and test splits. Unless otherwise noted, results are mean$\pm$std over 5 random seeds.

Baselines cover classic MIL (AB-MIL, CLAM-SB, DTFD-MIL), dual-stream and Transformer MIL (DS-MIL, TransMIL), and recent high-capacity methods including ILRA-MIL, MHIM-MIL, DGR-MIL, and 2DMambaMIL. Full setup details are in Appendix \ref{sec:setup}.

The evaluation asks three questions: whether the phenomenon can be isolated under controlled composition/topology, whether the residual design improves real WSI prediction under one feature pipeline, and whether the gains come with stronger spatial behavior rather than only higher capacity.

\subsection{Controlled Evidence for Spatial Blindness}
\label{sec:synthetic_main}

Spatial-MNIST-Bag is first used to separate composition from topology. Dataset A is purely compositional: coordinates are random and the label depends only on whether digit ``9'' appears. Dataset B removes this shortcut: every bag contains the same digit multiset, and the label depends only on whether five key digits form a compact spatial motif. Full construction details are in Appendix \ref{sec:spatial_mnist_bag}.

\begin{wraptable}[12]{r}{0.48\textwidth}
    \vspace{-6pt}
    \centering
    \scriptsize
    \caption{\textbf{Composition--Topology Diagnostic Benchmark.} Dataset A isolates composition; Dataset B requires topology.}
    \label{tab:synthetic_double_dissociation}
    \vspace{-5pt}
    \setlength{\tabcolsep}{3.2pt}
    \renewcommand{\arraystretch}{0.95}
    \resizebox{0.47\textwidth}{!}{
    \begin{tabular}{l cc}
        \toprule
        \textbf{Method} & \textbf{A: Comp.} & \textbf{B: Topo.} \\
        & \textbf{AUC} & \textbf{AUC} \\
        \midrule
        AB-MIL & \textbf{0.998} & 0.505 \\
        TransMIL & 0.995 & 0.532 \\
        ResTopoMIL (Joint) & 0.991 & 0.684 \\
        \midrule
        \textbf{ResTopoMIL} & 0.994 & \textbf{0.987} \\
        \bottomrule
    \end{tabular}}
    \vspace{-4pt}
\end{wraptable}

Table \ref{tab:synthetic_double_dissociation} gives the main diagnostic result. AB-MIL and TransMIL solve Dataset A, so their failure is not basic MIL capacity. The same models collapse on Dataset B, where composition is identical across classes. Joint training helps but remains far below residual decoupling. In this controlled setting, the issue is not an easier compositional rule; the statistical stream has no real label signal. The difficulty is that weak bag-level supervision must be assigned to a sparse spatial motif, while a jointly trained model can still fit non-informative statistical noise before the graph branch has learned the motif.
\FloatBarrier

\subsection{WSI Classification}

Table \ref{tab:classification_main} combines the four classification benchmarks and reports Accuracy/AUC; F1 scores, parameter counts, and CTransPath results are in Appendices \ref{sec:extra_cls_metrics} and \ref{sec:ctranspath}.

\begin{table}[H]
    \centering
    \small
    \vspace{-4pt}
    \caption{\textbf{Classification Results on Four WSI Benchmarks.} Accuracy/AUC are reported as mean $\pm$ std. F1 scores and parameter counts are separated into Appendix \ref{sec:extra_cls_metrics}.}
    \label{tab:classification_main}
    \setlength{\tabcolsep}{3.4pt}
    \renewcommand{\arraystretch}{1.03}
    \resizebox{\textwidth}{!}{
    \begin{tabular}{l cc cc cc cc}
        \toprule
        \multirow{2}{*}{\textbf{Method}} &
        \multicolumn{2}{c}{\textbf{BRACS}} &
        \multicolumn{2}{c}{\textbf{PANDA}} &
        \multicolumn{2}{c}{\textbf{TCGA-NSCLC}} &
        \multicolumn{2}{c}{\textbf{TCGA-BRCA}} \\
        \cmidrule(lr){2-3} \cmidrule(lr){4-5} \cmidrule(lr){6-7} \cmidrule(lr){8-9}
        & Acc & AUC & Acc & AUC & Acc & AUC & Acc & AUC \\
        \midrule
        AB-MIL &
        0.7275$_{\pm.0759}$ & 0.8806$_{\pm.0091}$ &
        0.7322$_{\pm.0059}$ & 0.9306$_{\pm.0017}$ &
        0.8988$_{\pm.0066}$ & 0.9569$_{\pm.0073}$ &
        0.9414$_{\pm.0062}$ & 0.9727$_{\pm.0019}$ \\
        CLAM-SB &
        \underline{0.7371}$_{\pm.0182}$ & \underline{0.8840}$_{\pm.0131}$ &
        0.7318$_{\pm.0041}$ & 0.9215$_{\pm.0009}$ &
        0.8836$_{\pm.0085}$ & 0.9635$_{\pm.0064}$ &
        0.9314$_{\pm.0062}$ & \underline{0.9814}$_{\pm.0027}$ \\
        DS-MIL &
        0.6460$_{\pm.0189}$ & 0.8054$_{\pm.0225}$ &
        0.7394$_{\pm.0207}$ & \underline{0.9309}$_{\pm.0070}$ &
        0.8836$_{\pm.0085}$ & 0.9579$_{\pm.0083}$ &
        0.9409$_{\pm.0124}$ & 0.9777$_{\pm.0039}$ \\
        TransMIL &
        0.6506$_{\pm.0174}$ & 0.8450$_{\pm.0098}$ &
        0.7090$_{\pm.0088}$ & 0.9288$_{\pm.0038}$ &
        0.8933$_{\pm.0132}$ & 0.9692$_{\pm.0084}$ &
        0.9409$_{\pm.0095}$ & 0.9787$_{\pm.0154}$ \\
        ILRA-MIL &
        0.6230$_{\pm.0286}$ & 0.8012$_{\pm.0219}$ &
        \underline{0.7402}$_{\pm.0096}$ & 0.9266$_{\pm.0034}$ &
        0.8912$_{\pm.0157}$ & 0.9628$_{\pm.0095}$ &
        0.9455$_{\pm.0051}$ & 0.9621$_{\pm.0148}$ \\
        MHIM-MIL &
        0.6690$_{\pm.0358}$ & 0.8340$_{\pm.0128}$ &
        0.6970$_{\pm.0123}$ & 0.9155$_{\pm.0020}$ &
        0.8908$_{\pm.0183}$ & \textbf{0.9759}$_{\pm.0030}$ &
        \underline{0.9465}$_{\pm.0095}$ & 0.9769$_{\pm.0129}$ \\
        DGR-MIL &
        0.7126$_{\pm.0363}$ & 0.8258$_{\pm.0386}$ &
        0.6964$_{\pm.0129}$ & 0.9043$_{\pm.0082}$ &
        0.9036$_{\pm.0225}$ & 0.9390$_{\pm.0224}$ &
        0.9455$_{\pm.0051}$ & 0.9724$_{\pm.0197}$ \\
        2DMambaMIL &
        0.7185$_{\pm.0290}$ & 0.8315$_{\pm.0310}$ &
        0.7012$_{\pm.0115}$ & 0.9105$_{\pm.0075}$ &
        \underline{0.9008}$_{\pm.0190}$ & 0.9450$_{\pm.0180}$ &
        0.9490$_{\pm.0060}$ & 0.9755$_{\pm.0180}$ \\
        \midrule
        \textbf{ResTopoMIL} &
        \textbf{0.7494}$_{\pm.0286}$ & \textbf{0.9006}$_{\pm.0055}$ &
        \textbf{0.7546}$_{\pm.0094}$ & \textbf{0.9426}$_{\pm.0010}$ &
        \textbf{0.9157}$_{\pm.0085}$ & \underline{0.9753}$_{\pm.0029}$ &
        \textbf{0.9568}$_{\pm.0095}$ & \textbf{0.9838}$_{\pm.0049}$ \\
        \bottomrule
    \end{tabular}
    }
    \vspace{-6pt}
\end{table}

ResTopoMIL gives the strongest overall classification profile in Table \ref{tab:classification_main}: it has the best Accuracy on all four datasets and the best AUC on three of four datasets, with MHIM-MIL slightly higher on TCGA-NSCLC AUC. The differences are most revealing on BRACS and PANDA, where atypia and Gleason grading depend strongly on tissue organization. On TCGA-BRCA the accuracy margins are smaller, but AUC and F1 remain consistent with the ductal-versus-lobular growth-pattern distinction.
\FloatBarrier

\subsection{WSI Survival Prediction}

Survival prediction is a useful stress test because labels are weaker and more heterogeneous than diagnostic labels. Table \ref{tab:result_survival_prediction} shows the best C-index on all five cohorts.

\begin{table}[H]
    \centering
    \small
    \vspace{-4pt}
    \caption{\textbf{Survival Prediction Results (C-Index).} Concordance Index is reported as mean $\pm$ std on 5 TCGA datasets. Parameter counts are listed separately in Appendix \ref{sec:extra_cls_metrics}. Higher C-Index is better.}
    \label{tab:result_survival_prediction}
    \setlength{\tabcolsep}{6pt}
    \renewcommand{\arraystretch}{1.04}
    \resizebox{\textwidth}{!}{
    \begin{tabular}{l c c c c c}
        \toprule
        \textbf{Method} & \textbf{KIRC} & \textbf{KIRP} & \textbf{LUAD} & \textbf{STAD} & \textbf{UCEC} \\
        \midrule
        AB-MIL &
        0.5694$_{\pm.0279}$ & 0.7091$_{\pm.0592}$ & 0.5942$_{\pm.0120}$ & 0.5871$_{\pm.0090}$ & 0.6220$_{\pm.0241}$ \\
        CLAM-SB &
        0.5705$_{\pm.0101}$ & 0.7034$_{\pm.0565}$ & 0.6192$_{\pm.0108}$ & 0.5829$_{\pm.0243}$ & 0.6034$_{\pm.0398}$ \\
        DS-MIL &
        0.5832$_{\pm.0207}$ & 0.6875$_{\pm.0456}$ & 0.5669$_{\pm.0145}$ & 0.6090$_{\pm.0187}$ & 0.6368$_{\pm.0417}$ \\
        TransMIL &
        0.5677$_{\pm.0162}$ & 0.7000$_{\pm.0238}$ & 0.6148$_{\pm.0124}$ & 0.6204$_{\pm.0359}$ & 0.6595$_{\pm.0486}$ \\
        ILRA-MIL &
        0.7276$_{\pm.0271}$ & 0.7272$_{\pm.1322}$ & 0.6067$_{\pm.0156}$ & 0.6589$_{\pm.0354}$ & 0.6767$_{\pm.0308}$ \\
        MHIM-MIL &
        0.7170$_{\pm.0143}$ & \underline{0.8073}$_{\pm.0545}$ & 0.5491$_{\pm.0205}$ & 0.6357$_{\pm.0321}$ & 0.6474$_{\pm.0704}$ \\
        DGR-MIL &
        0.7298$_{\pm.0068}$ & 0.8071$_{\pm.0265}$ & 0.6245$_{\pm.0117}$ & \underline{0.6625}$_{\pm.0122}$ & 0.6976$_{\pm.0104}$ \\
        2DMambaMIL &
        \underline{0.7311}$_{\pm.0110}$ & 0.8027$_{\pm.0250}$ & \underline{0.6290}$_{\pm.0095}$ & 0.6515$_{\pm.0150}$ & \underline{0.7020}$_{\pm.0140}$ \\
        \midrule
        \textbf{ResTopoMIL} &
        \textbf{0.7313}$_{\pm.0104}$ & \textbf{0.8182}$_{\pm.0227}$ & \textbf{0.6457}$_{\pm.0070}$ & \textbf{0.6807}$_{\pm.0086}$ & \textbf{0.7058}$_{\pm.0128}$ \\
        \bottomrule
    \end{tabular}
    }
    \vspace{-10pt}
\end{table}

The advantage is clearest on LUAD and STAD, where growth patterns and local tissue organization are prognostic factors. The consistent C-index gains suggest that the residual branch captures more than tissue proportions.
\FloatBarrier

\subsection{Mechanistic and Ablation Evidence}

\begin{figure}[tbp]
    \centering
    \includegraphics[width=0.68\textwidth]{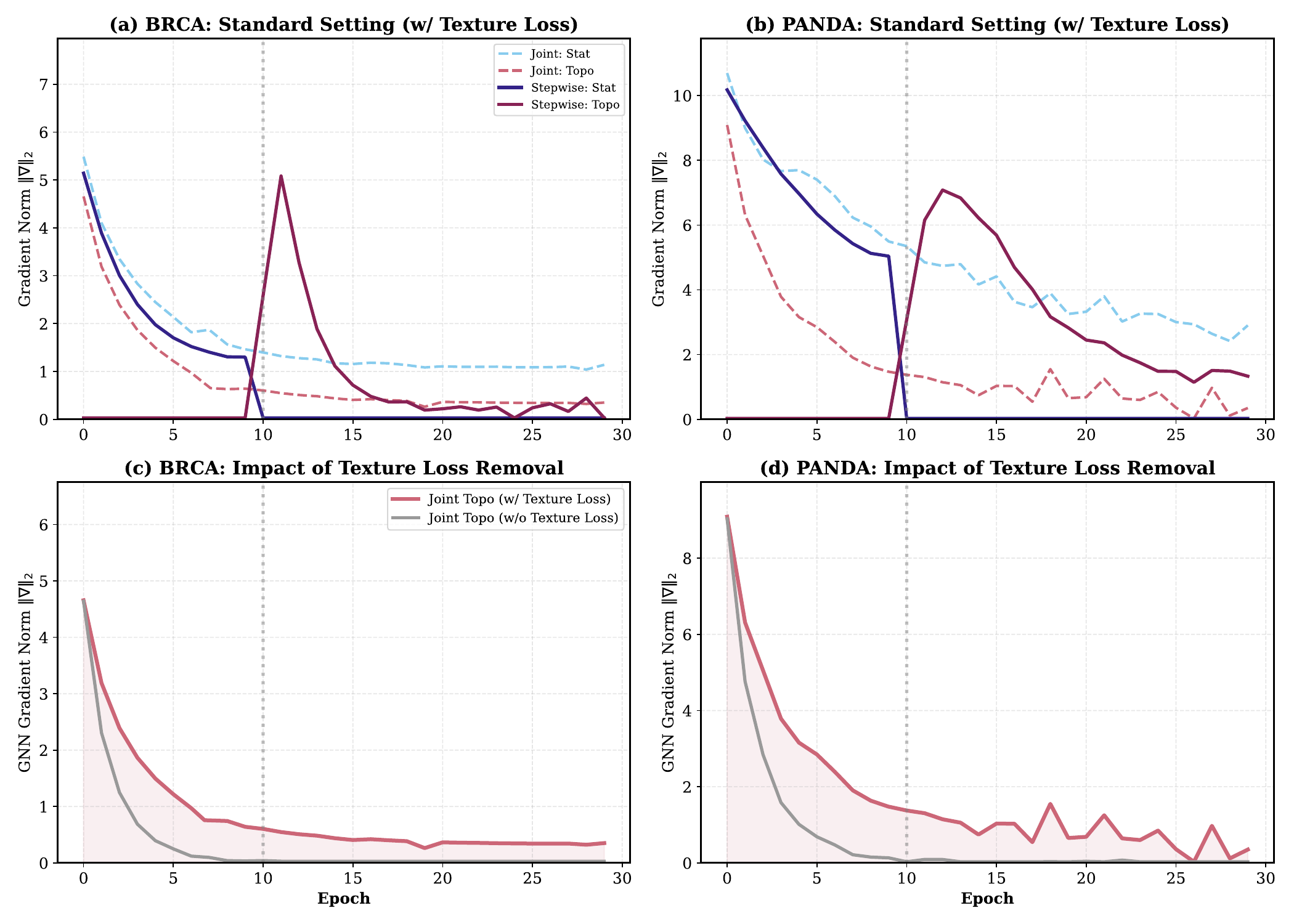}
    \caption{\textbf{Gradient dynamics.} Stepwise training revives the topological gradient after freezing the statistical stream; joint optimization and the variant without $\mathcal{L}_{texture}$ both let it fade.}
    \label{fig:gradient}
    \vspace{-12pt}
\end{figure}

The mechanistic question is whether the gain comes from residual decoupling rather than a larger backbone, a training trick, or a favorable seed. Figure \ref{fig:gradient} and Tables \ref{tab:core_strategy_ablation}--\ref{tab:arch_validity_ablation} give the main evidence; Appendix \ref{sec:ablation_table} gives full metrics.

Figure \ref{fig:gradient} separates endpoint performance from training dynamics. In the joint run, slide-level AUC improves early and then saturates while the graph branch receives little gradient. In the stepwise run, unfreezing the graph branch produces a second rise, consistent with residual errors becoming available to topology after the statistical stream is fixed. The gradient trace is the sharper evidence: the topological gradient fades under joint optimization, rebounds after freezing, and collapses again when $\mathcal{L}_{texture}$ is removed. Thus the same GCN needs both a fixed residual target and a coordinate-specific constraint.

\begin{table}[H]
    \centering
    \scriptsize
    \setlength{\tabcolsep}{1.6pt}
    \renewcommand{\arraystretch}{0.86}
    \begin{minipage}[t]{0.49\textwidth}
        \centering
        \caption{\textbf{Core Strategy.} AUC mean$\pm$std.}
        \label{tab:core_strategy_ablation}
        \resizebox{\linewidth}{!}{
        \begin{tabular}{llcc}
            \toprule
            \multirow{2}{*}{\textbf{Group}} & \multirow{2}{*}{\textbf{Variant}} &
            \textbf{PANDA} &
            \textbf{BRCA} \\
            & & AUC & AUC \\
            \midrule
            \multirow{5}{*}{\shortstack[l]{Opt.}}
            & \textbf{ResTopoMIL} & \textbf{0.9426}$_{\pm.0010}$ & 0.9838$_{\pm.0049}$ \\
            & Stat. Only & 0.9027$_{\pm.0051}$ & 0.9486$_{\pm.0035}$ \\
            & Topo. Only & 0.9215$_{\pm.0055}$ & 0.9608$_{\pm.0048}$ \\
            & Joint Opt. & 0.9299$_{\pm.0052}$ & 0.9773$_{\pm.0022}$ \\
            & Multi-LR & 0.9352$_{\pm.0048}$ & 0.9786$_{\pm.0018}$ \\
            \midrule
            \multirow{3}{*}{Fusion}
            & Gated Fusion & 0.9225$_{\pm.0062}$ & 0.9802$_{\pm.0042}$ \\
            & MoE & 0.9305$_{\pm.0035}$ & \textbf{0.9897}$_{\pm.0028}$ \\
            & Indep. Clf & 0.9256$_{\pm.0062}$ & 0.9845$_{\pm.0025}$ \\
            \midrule
            Constraint & w/o $\mathcal{L}_{tex}$ & 0.9147$_{\pm.0065}$ & 0.9762$_{\pm.0038}$ \\
            \bottomrule
        \end{tabular}}
    \end{minipage}
    \hfill
    \begin{minipage}[t]{0.49\textwidth}
        \centering
        \caption{\textbf{Architecture \& Validity.} AUC mean$\pm$std.}
        \label{tab:arch_validity_ablation}
        \resizebox{\linewidth}{!}{
        \begin{tabular}{llcc}
            \toprule
            \multirow{2}{*}{\textbf{Group}} & \multirow{2}{*}{\textbf{Variant}} &
            \textbf{PANDA} &
            \textbf{BRCA} \\
            & & AUC & AUC \\
            \midrule
            \multirow{2}{*}{Backbone}
            & \textbf{ResTopoMIL (GCN)} & \textbf{0.9426}$_{\pm.0010}$ & 0.9838$_{\pm.0049}$ \\
            & GAT & 0.9419$_{\pm.0041}$ & \textbf{0.9889}$_{\pm.0025}$ \\
            \midrule
            \multirow{2}{*}{Sanity}
            & Random Graph & 0.8286$_{\pm.0150}$ & 0.8563$_{\pm.0110}$ \\
            & Fixed Proto. & 0.9349$_{\pm.0055}$ & 0.9746$_{\pm.0035}$ \\
            \bottomrule
        \end{tabular}}
    \end{minipage}
    \vspace{-12pt}
\end{table}

The compact tables are best read together rather than as a single leaderboard. Stat.-Only and Topo.-Only show that neither branch alone explains the full behavior: composition is a strong baseline, but topology has independent signal. Joint Opt. and Multi-LR are the closest alternatives to our schedule; the latter helps on PANDA, but still does not match residual decoupling. MoE and GAT reach higher TCGA-BRCA AUC than the default GCN setting, yet they are weaker on PANDA and do not give the same controlled spatial behavior seen in the coordinate-shuffling diagnostic. The sanity controls are more decisive: a random graph breaks performance, fixed prototypes weaken the anchor, and removing $\mathcal{L}_{tex}$ is the strongest non-random negative control on PANDA AUC.

These results are not meant to argue that a two-layer GCN is intrinsically better than attention or state-space context modules. The point is narrower: once a strong compositional shortcut is available, a spatial module can be present but weakly used. Residual decoupling changes the assignment of error. It lets the statistical stream explain the easy part, then gives the remaining supervision to a branch explicitly tested against coordinate corruption. This is also why high slide-level AUC should not be taken as proof that a model has used tissue architecture; the model should predict well, but lose the right evidence when spatial structure is deliberately removed.

\section{Conclusion}

This paper identifies \textit{spatial blindness}: context-aware MIL can perform well at slide level while using little tissue topology. ResTopoMIL addresses this failure mode by fitting composition first and then training topology as a residual correction under a shuffle-based spatial constraint. Across 9 public benchmarks it improves prediction, restores spatial sensitivity, and gives stronger localization evidence with only 1.15M parameters. The broader lesson is that adding context is not the same as learning from context; optimization can decide which signal is actually used. Future work should test this diagnosis with trainable pathology foundation encoders, prospective cohorts, and clinically realistic coordinate perturbations such as registration noise or rigid transformations.

\clearpage
\bibliographystyle{plainnat}
\bibliography{neurips2026_qill0506}

\clearpage
\appendix

\section{Overview}
\label{sec:appendix_overview}

This supplement provides the full technical and experimental material behind the main paper. Appendix \ref{sec:theory} presents the design analysis of residual decoupling, including the residual-error view of graph gradients and the conditional-information interpretation of the topological branch. Appendix \ref{sec:setup} gives the experimental configuration, covering the WSI datasets, survival cohorts, Spatial-MNIST-Bag construction, patch processing, implementation details, and training protocol. Appendix \ref{sec:extra_cls_metrics} reports additional classification metrics and model sizes. Appendix \ref{sec:ablation_table} provides the extended ablation analysis for optimization strategy, fusion design, graph validity, and hyperparameter sensitivity. Appendix \ref{sec:ctranspath} evaluates cross-backbone generalization with CTransPath features. Appendix \ref{sec:shuffle_four_benchmarks} reports additional shuffle sensitivity on WSI benchmarks, and Appendix \ref{sec:shuffle_appendix} gives the progressive coordinate-shuffling analysis. Appendix \ref{sec:camelyon_localization} details the CAMELYON-16 localization protocol and quantitative localization results. Appendix \ref{sec:feature_space_vis} visualizes the feature spaces learned by the statistical and topological streams. Appendix \ref{sec:vis} provides qualitative heatmap analysis, and Appendix \ref{sec:additional_limitations} discusses additional limitations and negative-result scope.

\section{Design Analysis of Residual Decoupling}
\label{sec:theory}

This appendix derives the two propositions used in Section \ref{sec:theoretical_insights}. The scope is deliberately narrow. We do not claim that every MIL model must suffer from gradient starvation, or that conditional mutual information is new. The aim is to spell out what the stop-gradient residual objective does in ResTopoMIL: it separates a compositional anchor from a spatial correction.

\subsection{Residual-Error Gating of the Graph Gradient}
\label{sec:proof_grad_starvation}

For binary classification, ResTopoMIL uses an additive logit
\begin{equation}
    f(X)=f_{stat}(X;\theta_s)+f_{topo}(X;\theta_t),
    \qquad
    \hat p_\theta(X)=P_\theta(Y=1\mid X)=\sigma(f(X)).
\end{equation}
With cross-entropy loss
\begin{equation}
    \mathcal{L}
    =
    \mathbb{E}\big[-Y\log \hat p_\theta(X)-(1-Y)\log(1-\hat p_\theta(X))\big],
\end{equation}
the gradient received by the topological branch is
\begin{equation}
    \nabla_{\theta_t}\mathcal{L}
    =
    \mathbb{E}\big[(\hat p_\theta(X)-Y)\nabla_{\theta_t}f_{topo}(X;\theta_t)\big].
    \label{eq:exact_topo_gradient}
\end{equation}
Let $r_\theta(X,Y)=\hat p_\theta(X)-Y$ denote the remaining prediction error. Applying Cauchy--Schwarz to Eq. \eqref{eq:exact_topo_gradient} gives
\begin{equation}
    \|\nabla_{\theta_t}\mathcal{L}\|
    \leq
    \left(\mathbb{E}r_\theta(X,Y)^2\right)^{1/2}
    \left(\mathbb{E}\|\nabla_{\theta_t}f_{topo}(X;\theta_t)\|_F^2\right)^{1/2}.
    \label{eq:topo_grad_bound}
\end{equation}
The bound gives the method-specific point: the graph update is gated by the residual error of the current additive predictor. If the statistical stream rapidly explains much of the slide label, the magnitude of $r_\theta(X,Y)$ becomes small before the graph branch has learned a useful spatial rule. The graph module may be present in the architecture but weakly trained in practice.

\paragraph{Why the stop-gradient matters.}
Stage 2 replaces the moving joint target with
\begin{equation}
    f(X)=\sg[f_{stat}(X)]+f_{topo}(X;\theta_t).
\end{equation}
The residual error is now computed against a fixed compositional anchor. The statistical stream can no longer reduce this residual during Stage 2, so the remaining errors are assigned to the topological branch. This is the specific optimization change in ResTopoMIL. Figure \ref{fig:gradient} is the empirical counterpart: the graph gradient rebounds after unfreezing in the stepwise run, while it decays under joint training. The ablations give additional checks: GAT does not rescue joint training, Multi-LR improves only modestly, and removing $\mathcal{L}_{tex}$ again weakens the graph branch.

\subsection{What Information the Residual Branch Is Asked to Learn}
\label{sec:proof_mi_residual}

Let $Z_{stat}=g_s(X)$ and $Z_{topo}=g_t(X)$ denote the statistical and topological representations. The information that the two streams jointly provide about the label decomposes by the chain rule:
\begin{equation}
    I(Z_{stat},Z_{topo};Y)
    =
    I(Z_{stat};Y)
    +
    I(Z_{topo};Y\mid Z_{stat}).
    \label{eq:mi_chain}
\end{equation}
The first term corresponds to label information already captured by the statistical anchor. The second term is the extra information that remains after conditioning on that anchor. ResTopoMIL is designed around this second term, but with an important practical detail: the topology representation is trained through logits added to a frozen statistical predictor.

In Stage 1, the statistical branch is optimized to obtain $Z_{stat}$ and a statistical logit $f_{stat}$. In Stage 2, $Z_{stat}$ and $f_{stat}$ are frozen, and the topological branch parameterizes a variational conditional posterior
\begin{equation}
    q_{\phi}(Y=y\mid Z_{stat},Z_{topo})
    =
    \left[\mathrm{Softmax}\!\big(f_{stat}(Z_{stat})+f_{topo}(Z_{topo};\phi)\big)\right]_y .
\end{equation}
For binary classification, this reduces to the sigmoid probability of the positive class, $q_{\phi}(Y=1\mid Z_{stat},Z_{topo})=\sigma(f_{stat}+f_{topo})$.
The conditional mutual information can be written as
\begin{equation}
    I(Z_{topo};Y\mid Z_{stat})
    = H(Y\mid Z_{stat})-H(Y\mid Z_{stat},Z_{topo}).
\end{equation}
Since $Z_{stat}$ is fixed in Stage 2, $H(Y\mid Z_{stat})$ is constant with respect to $\phi$. It remains to minimize the second term. For any variational decoder $q_{\phi}$,
\begin{align}
    H(Y\mid Z_{stat},Z_{topo})
    &= \mathbb{E}\big[-\log p(Y\mid Z_{stat},Z_{topo})\big] \\
    &\leq \mathbb{E}\big[-\log q_{\phi}(Y\mid Z_{stat},Z_{topo})\big],
\end{align}
where the inequality follows from the non-negativity of
\begin{equation}
    \mathbb{E}_{Z_{stat},Z_{topo}}
    \mathrm{KL}\big(p(Y\mid Z_{stat},Z_{topo})\|q_{\phi}(Y\mid Z_{stat},Z_{topo})\big).
\end{equation}
Therefore,
\begin{equation}
    I(Z_{topo};Y\mid Z_{stat})
    \geq
    H(Y\mid Z_{stat})
    +
    \mathbb{E}\big[\log q_{\phi}(Y\mid Z_{stat},Z_{topo})\big].
    \label{eq:cmi_lower_bound}
\end{equation}
Maximizing the right-hand side is equivalent, up to the fixed constant $H(Y\mid Z_{stat})$, to minimizing the Stage-2 cross-entropy loss
\begin{equation}
    \mathcal{L}_{res}
    =
    \mathbb{E}\left[
    -\log q_{\phi}(Y\mid Z_{stat},Z_{topo})
    \right].
\end{equation}
Thus the residual objective maximizes a variational lower bound on $I(Z_{topo};Y\mid Z_{stat})$ with the statistical stream held fixed. This does not claim that the learned representation is automatically spatial. It only states what label information the residual decoder is optimized to extract once the anchor is frozen.

\paragraph{Why conditional information is not enough.}
The chain rule alone would still allow $Z_{topo}$ to encode another compositional statistic not captured by $Z_{stat}$. ResTopoMIL therefore adds a coordinate-specific constraint. In the shuffled view, the patch multiset and slide label are unchanged, but the KNN graph induced by physical coordinates is corrupted. A residual branch that ignores coordinates will map the clean and shuffled views to similar representations and will be penalized by $\mathcal{L}_{texture}$. This is the part of the method that turns a generic residual objective into a topology-seeking residual objective.

\subsection{Why Common Optimization Heuristics Are Not Equivalent}
\label{sec:theory_optim_heuristics}

The residual view also clarifies why simple training tricks help only partially. Let
\begin{equation}
    r_{\theta_s,\theta_t}(X,Y)=\hat p_{\theta_s,\theta_t}(X)-Y
\end{equation}
be the error term multiplying the graph gradient in Eq. \eqref{eq:exact_topo_gradient}. In joint optimization, both streams reduce the same residual:
\begin{equation}
    \mathbb{E}[\Delta\theta_t]
    =
    -\eta_t\,\mathbb{E}\!\left[
    r_{\theta_s,\theta_t}(X,Y)\nabla_{\theta_t}f_{topo}(X)
    \right].
    \label{eq:joint_update}
\end{equation}
If the statistical stream quickly reduces the magnitude of $r_{\theta_s,\theta_t}$, the update in Eq. \eqref{eq:joint_update} becomes small even when the graph branch has not learned a meaningful spatial rule. This is the optimization form of spatial blindness.

A larger graph learning rate changes Eq. \eqref{eq:joint_update} only by a scalar. With a multiplier $\alpha>1$,
\begin{equation}
    \mathbb{E}[\Delta\theta_t]
    =
    -\alpha\eta_t\,\mathbb{E}\!\left[
    r_{\theta_s,\theta_t}(X,Y)\nabla_{\theta_t}f_{topo}(X)
    \right].
\end{equation}
The graph branch moves faster while the residual is nonzero, but the update is still gated by the same shrinking residual. This explains why Multi-LR improves over vanilla joint training in Table \ref{tab:app_opt_ablation}, but remains below the frozen-anchor schedule.

Statistical dropout attacks the same problem from the opposite side. By corrupting the statistical stream, it can increase the residual visible to the graph branch:
\begin{equation}
    r_{\mathrm{drop}}(X,Y)=\hat p(Y=1\mid D(f_{stat}),f_{topo})-Y ,
\end{equation}
where $D(\cdot)$ denotes dropout. This may expose topology, but it also injects noise into useful compositional evidence. The higher variance of the Stat-Dropout row is consistent with this trade-off. Curriculum scheduling is a softer decoupling: it changes the relative speed of the two streams, but unless the statistical branch is frozen, the residual target remains moving. Hard instance mining reweights samples,
\begin{equation}
    \mathbb{E}[\Delta\theta_t]
    =
    -\eta_t\,\mathbb{E}\!\left[
    w(X) r_{\theta_s,\theta_t}(X,Y)\nabla_{\theta_t}f_{topo}(X)
    \right],
\end{equation}
and can help by emphasizing difficult slides; it still does not prevent the easy stream from absorbing part of the residual.

ResTopoMIL makes a stronger intervention:
\begin{equation}
    r_{\mathrm{res}}(X,Y)=\hat p(Y=1\mid \sg[f_{stat}(X)], f_{topo}(X))-Y,
\end{equation}
so the residual is measured against a fixed compositional anchor. The graph branch no longer competes with a moving statistical predictor for the same error term. The shuffle loss then restricts the residual representation to information that changes when the coordinate-induced graph is damaged. This is why the method should be read as an optimization design rather than as a more complicated graph architecture.

\section{Experimental Configuration}
\label{sec:setup}

\subsection{Datasets}
Evaluation was conducted on nine datasets spanning three tasks. \textbf{TCGA-NSCLC} contains 1,053 WSIs (LUAD: 541, LUSC: 512) for non-small cell lung cancer subtyping. \textbf{TCGA-BRCA} includes 1,021 WSIs distinguishing Invasive Ductal Carcinoma (IDC) from Invasive Lobular Carcinoma (ILC), with structural differences (ductal vs. single-file strands) serving as key discriminators. \textbf{BRACS} comprises 525 WSIs covering 7 fine-grained categories, where differentiation of atypical classes (ADH, FEA) relies on subtle microenvironmental changes. Finally, \textbf{PANDA} contains 10,616 WSIs for Gleason grading, a task highly dependent on structure (e.g., gland formation vs. fused glands) and serving as a primary topological benchmark. All WSI splits are stratified at the patient level whenever patient identifiers are available; slides from the same patient are never shared across train, validation, and test partitions.

\subsubsection{Survival Prediction}
Prognosis performance was evaluated on five TCGA cohorts selected for their structural prognostic factors. \textbf{TCGA-LUAD} (516 WSIs) prognosis correlates with histological growth patterns (lepidic, acinar, papillary, micropapillary, solid). \textbf{TCGA-STAD} (441 WSIs) utilizes differentiation grade and structural anomalies like signet-ring cells as indicators. \textbf{TCGA-UCEC} (537 WSIs) prognosis links to FIGO grade, defined by the glandular-to-solid component ratio. \textbf{TCGA-KIRC} (519 WSIs) survival outcomes associate with Fuhrman grade (nuclear characteristics within clear cell architecture), while \textbf{TCGA-KIRP} (259 WSIs) depends on the structural integrity and arrangement of papillary cores. The task formulation is discrete time-to-event prediction across 4 non-overlapping intervals.

\subsection{Spatial-MNIST-Bag Construction}
\label{sec:spatial_mnist_bag}

To rigorously separate statistical composition from spatial topology, a controlled synthetic benchmark is constructed from MNIST, termed \textbf{Spatial-MNIST-Bag}. Raw MNIST images are first encoded into 512-dimensional embeddings using a pre-trained ResNet-18. Each bag is represented as
\begin{equation}
    X=\{(h_i,p_i)\}_{i=1}^{N},\qquad N=50,\qquad p_i\in[0,1]^2,
\end{equation}
where $h_i$ is the digit embedding and $p_i$ is its assigned two-dimensional coordinate. The benchmark contains two complementary datasets.

\paragraph{Dataset A: Pure Composition.}
Dataset A follows the standard i.i.d. MIL assumption and evaluates whether a model can detect a statistical anomaly without relying on structural cues. For every bag, all coordinates are sampled independently from $U(0,1)\times U(0,1)$, so the spatial graph is random noise. The bag label depends only on digit composition: $Y=1$ if at least one instance is digit ``9'', and $Y=0$ otherwise. A successful model should behave as a permutation-invariant instance detector.

\paragraph{Dataset B: Pure Topology.}
Dataset B removes compositional shortcuts and forces the model to use spatial reasoning. Every bag contains the same predefined digit multiset: exactly one ``1'', one ``3'', one ``5'', one ``7'', one ``9'', and 45 even digits sampled from $\{0,2,4,6,8\}$. Therefore, a purely statistical MIL model observes the same bag-level composition for positive and negative samples and should remain close to chance.

The label in Dataset B is determined only by the coordinates of the five key odd digits. For positive bags ($Y=1$), a centroid $c\sim U([0.2,0.8]^2)$ is sampled and the coordinates of digits ``1'', ``3'', ``5'', ``7'', and ``9'' are drawn from $\mathcal{N}(c,\sigma^2 I)$ with $\sigma=0.05$, creating a compact topological motif that forms a dense clique in the KNN graph. The remaining 45 background digits are uniformly scattered in $[0,1]^2$. For negative bags ($Y=0$), all 50 instances, including the five key odd digits, are sampled uniformly in $[0,1]^2$ with no spatial correlation. The task is therefore impossible to solve from composition alone and directly tests whether a model can detect a non-i.i.d. spatial motif.

\begin{table}[H]
    \centering
    \small
    \caption{\textbf{Spatial-MNIST-Bag Generation Parameters.} Dataset A isolates composition, while Dataset B neutralizes composition and makes the label depend only on a spatial motif.}
    \label{tab:spatial_mnist_params}
    \setlength{\tabcolsep}{5pt}
    \renewcommand{\arraystretch}{1.08}
    \resizebox{\textwidth}{!}{
    \begin{tabular}{lcc}
        \toprule
        \textbf{Item} & \textbf{Dataset A: Pure Composition} & \textbf{Dataset B: Pure Topology} \\
        \midrule
        Instance feature & \multicolumn{2}{c}{512-d ResNet-18 embedding of raw MNIST digits} \\
        Bag size & \multicolumn{2}{c}{$N=50$ instances} \\
        Coordinate domain & \multicolumn{2}{c}{$p_i\in[0,1]^2$} \\
        Composition & Random MNIST digits; positive if digit ``9'' appears &
        Fixed multiset: one each of ``1,3,5,7,9'' plus 45 even digits \\
        Positive rule & At least one digit ``9'' in the bag &
        Key odd digits form a compact cluster \\
        Negative rule & No digit ``9'' in the bag &
        All instances, including key digits, are uniformly scattered \\
        Key motif & None; coordinates are nuisance noise &
        Centroid $c\sim U([0.2,0.8]^2)$, key coordinates $\sim\mathcal{N}(c,0.05^2I)$ \\
        Required capability & Permutation-invariant instance detection & Detection of a non-i.i.d. spatial motif \\
        \bottomrule
    \end{tabular}
    }
\end{table}

This construction removes a common ambiguity in real WSI data. Dataset A checks whether the model still behaves like a standard MIL classifier when topology is irrelevant. Dataset B keeps the instance composition fixed and makes spatial arrangement the only usable signal. The two settings separate questions that are usually entangled in pathology benchmarks: whether a method recognizes discriminative patches, and whether it uses the physical arrangement of those patches once composition is no longer enough.

\subsection{Preprocessing and Feature Extraction}
Data processing followed standard MIL protocols (e.g., CLAM, TransMIL). Tissue extraction was performed via Otsu's thresholding in HSV space, followed by the generation of non-overlapping $256 \times 256$ patches at 20$\times$ magnification. Feature extraction used the \textbf{UNI} foundation model, a ViT-Large architecture pre-trained via self-supervised learning (DINOv2) on Mass-100K ($>10^8$ patches). This yields robust 1024-dimensional embeddings; all baselines used identical UNI features.

\subsection{Implementation Details}
Experiments were conducted using PyTorch 1.13 and PyTorch Geometric on a single NVIDIA RTX 3090 (24GB).

\paragraph{Architecture.} The statistical stream utilizes a codebook size of $K=32$ with a learnable temperature $\tau$ (init 1.0). Prototype initialization uses MiniBatch K-Means on a random subset of training patch embeddings, which makes the initialization scalable to WSI collections with millions of patches. The topological stream employs a KNN graph ($K_{knn}=8$) processed by a 2-layer GCN with ReLU and Dropout ($p=0.25$). Both streams map to a 512-dim hidden space, and node embeddings are aggregated by global mean pooling before the residual classifier.

\paragraph{Training Protocol.} A decoupled two-stage strategy was employed. Stage 1 (Warmup) optimizes the Statistical Stream for 10 epochs. Stage 2 (Refinement) freezes the Statistical Stream and optimizes the Topological Stream (30 epochs for classification, 20 for survival). Optimization used Lookahead (Adam base, weight decay $1\times10^{-4}$) with a Cosine Annealing scheduler (LR $2\times10^{-4}$). Loss scaling factors were set to margin $m=0.3$ and weight $\lambda=1.0$.

\section{Additional Classification Metrics and Model Size}
\label{sec:extra_cls_metrics}

The main text reports Accuracy and AUC for compactness. Table \ref{tab:app_params} lists the model sizes used in the same comparison, and Table \ref{tab:app_f1_cls} reports the corresponding F1 scores on the four classification datasets.

\begin{table}[H]
    \centering
    \small
    \caption{\textbf{Parameter Counts.} Model sizes are reported in millions of trainable parameters under the unified implementation.}
    \label{tab:app_params}
    \setlength{\tabcolsep}{7pt}
    \renewcommand{\arraystretch}{1.08}
    \begin{tabular}{lc}
        \toprule
        \textbf{Method} & \textbf{Params (M)} \\
        \midrule
        AB-MIL & 0.59 \\
        CLAM-SB & 0.79 \\
        DS-MIL & 1.20 \\
        TransMIL & 2.67 \\
        ILRA-MIL & 3.68 \\
        MHIM-MIL & 2.67 \\
        DGR-MIL & 4.35 \\
        2DMambaMIL & 1.27 \\
        \textbf{ResTopoMIL} & 1.15 \\
        \bottomrule
    \end{tabular}
\end{table}

The parameter comparison rules out a simple capacity explanation. ResTopoMIL has 1.15M trainable parameters, close to DS-MIL and smaller than TransMIL, ILRA-MIL, MHIM-MIL, DGR-MIL, and 2DMambaMIL under the same implementation. The gains in the main tables therefore do not come from a larger model budget, but from how the statistical and topological streams are trained and constrained.

\begin{table}[H]
    \centering
    \small
    \caption{\textbf{F1 Scores on Four Classification Benchmarks.} Mean $\pm$ std over 5 random seeds.}
    \label{tab:app_f1_cls}
    \setlength{\tabcolsep}{5pt}
    \renewcommand{\arraystretch}{1.08}
    \resizebox{\textwidth}{!}{
    \begin{tabular}{lcccc}
        \toprule
        \textbf{Method} & \textbf{BRACS} & \textbf{PANDA} & \textbf{TCGA-NSCLC} & \textbf{TCGA-BRCA} \\
        \midrule
        AB-MIL & 0.6748$_{\pm.0185}$ & 0.6876$_{\pm.0079}$ & 0.8680$_{\pm.0068}$ & 0.9187$_{\pm.0106}$ \\
        CLAM-SB & \underline{0.7051}$_{\pm.0287}$ & 0.6814$_{\pm.0050}$ & 0.8529$_{\pm.0084}$ & \underline{0.9268}$_{\pm.0112}$ \\
        DS-MIL & 0.6068$_{\pm.0436}$ & \underline{0.6911}$_{\pm.0231}$ & 0.8627$_{\pm.0099}$ & 0.9058$_{\pm.0208}$ \\
        TransMIL & 0.5592$_{\pm.0398}$ & 0.6667$_{\pm.0133}$ & 0.8822$_{\pm.0132}$ & 0.9016$_{\pm.0163}$ \\
        ILRA-MIL & 0.5526$_{\pm.0368}$ & 0.6880$_{\pm.0094}$ & 0.8829$_{\pm.0159}$ & 0.9108$_{\pm.0091}$ \\
        MHIM-MIL & 0.5874$_{\pm.0508}$ & 0.6372$_{\pm.0158}$ & 0.8703$_{\pm.0183}$ & 0.9117$_{\pm.0155}$ \\
        DGR-MIL & 0.6684$_{\pm.0509}$ & 0.6399$_{\pm.0178}$ & 0.8829$_{\pm.0225}$ & 0.9099$_{\pm.0098}$ \\
        2DMambaMIL & 0.6710$_{\pm.0450}$ & 0.6450$_{\pm.0160}$ & \underline{0.8855}$_{\pm.0210}$ & 0.9150$_{\pm.0110}$ \\
        \midrule
        \textbf{ResTopoMIL} & \textbf{0.7142}$_{\pm.0420}$ & \textbf{0.7097}$_{\pm.0098}$ & \textbf{0.9135}$_{\pm.0087}$ & \textbf{0.9308}$_{\pm.0166}$ \\
        \bottomrule
    \end{tabular}
    }
\end{table}

The F1 scores are consistent with Accuracy and AUC. ResTopoMIL obtains the best F1 on all four classification datasets, including fine-grained BRACS and structure-heavy PANDA. Since F1 is more sensitive to class imbalance and minority-class errors than accuracy, the improvement is not only a ranking effect in AUC; it also reflects more balanced classification behavior.

\section{Extended Ablation Analysis}
\label{sec:ablation_table}

The main text reports compact AUC-only ablations. Here we provide the full Accuracy/F1/AUC results with standard deviations, but split them into smaller tables so that each block answers a specific question.

\begin{table}[H]
    \centering
    \small
    \caption{\textbf{Optimization Ablation.} Full metrics for residual decoupling and softer optimization alternatives.}
    \label{tab:app_opt_ablation}
    \setlength{\tabcolsep}{4.2pt}
    \renewcommand{\arraystretch}{1.06}
    \resizebox{\textwidth}{!}{
    \begin{tabular}{l ccc ccc}
        \toprule
        \multirow{2}{*}{\textbf{Variant}} & \multicolumn{3}{c}{\textbf{PANDA}} & \multicolumn{3}{c}{\textbf{TCGA-BRCA}} \\
        \cmidrule(lr){2-4}\cmidrule(lr){5-7}
        & Acc & F1 & AUC & Acc & F1 & AUC \\
        \midrule
        Stat. Only & 0.6761$_{\pm.0062}$ & 0.6113$_{\pm.0075}$ & 0.9027$_{\pm.0051}$ & 0.9138$_{\pm.0069}$ & 0.8896$_{\pm.0112}$ & 0.9486$_{\pm.0035}$ \\
        Topo. Only & 0.7042$_{\pm.0072}$ & 0.6598$_{\pm.0085}$ & 0.9215$_{\pm.0055}$ & 0.9382$_{\pm.0068}$ & 0.8945$_{\pm.0102}$ & 0.9608$_{\pm.0048}$ \\
        Joint Opt. & 0.7397$_{\pm.0055}$ & 0.7025$_{\pm.0065}$ & 0.9299$_{\pm.0052}$ & 0.9432$_{\pm.0055}$ & 0.9318$_{\pm.0061}$ & 0.9773$_{\pm.0022}$ \\
        + Stat-Dropout (PDL) & 0.7415$_{\pm.0070}$ & 0.7038$_{\pm.0075}$ & 0.9325$_{\pm.0065}$ & 0.9448$_{\pm.0048}$ & 0.9302$_{\pm.0068}$ & 0.9780$_{\pm.0030}$ \\
        + Curriculum Sched. & 0.7442$_{\pm.0052}$ & 0.7055$_{\pm.0058}$ & 0.9348$_{\pm.0045}$ & 0.9465$_{\pm.0045}$ & 0.9315$_{\pm.0055}$ & 0.9785$_{\pm.0018}$ \\
        + Hard Instance Mining & 0.7485$_{\pm.0062}$ & 0.7091$_{\pm.0065}$ & 0.9372$_{\pm.0050}$ & 0.9492$_{\pm.0052}$ & 0.9288$_{\pm.0065}$ & 0.9792$_{\pm.0021}$ \\
        Joint Opt. (Multi-LR) & 0.7468$_{\pm.0051}$ & 0.7084$_{\pm.0045}$ & 0.9352$_{\pm.0048}$ & 0.9475$_{\pm.0042}$ & 0.9235$_{\pm.0054}$ & 0.9786$_{\pm.0018}$ \\
        \midrule
        \textbf{ResTopoMIL} & \textbf{0.7546}$_{\pm.0094}$ & \textbf{0.7097}$_{\pm.0098}$ & \textbf{0.9426}$_{\pm.0010}$ & \textbf{0.9568}$_{\pm.0095}$ & \textbf{0.9308}$_{\pm.0166}$ & \textbf{0.9838}$_{\pm.0049}$ \\
        \bottomrule
    \end{tabular}
    }
\end{table}

Table \ref{tab:app_opt_ablation} asks whether easier training heuristics can fix spatial blindness. Topo. Only is not a replacement for the statistical stream: it is close to Stat. Only on PANDA and higher on TCGA-BRCA, but does not provide the same balanced behavior as the residually decoupled model. The graph branch should model the structural residual left by a frozen compositional predictor, not solve the whole task alone. Stat-Dropout, curriculum scheduling, hard instance mining, and Multi-LR improve over vanilla joint training in several columns, but they are softer interventions: they rescale, perturb, schedule, or reweight the joint residual, whereas ResTopoMIL fixes the compositional anchor and changes what error the graph branch is asked to explain. The benefit is clearest when these endpoint metrics are read together with the shuffle and localization analyses, which test whether high AUC is accompanied by coordinate-dependent behavior.

\subsection{Training Dynamics and Computational Cost}
\label{sec:training_dynamics}

The two-stage schedule introduces a real training constraint, so it should not be treated as free. In our implementation, Stage 1 trains the statistical stream for 10 epochs. Stage 2 then freezes this stream and trains the graph branch for 30 epochs on classification tasks and 20 epochs on survival tasks. Compared with a single joint run using the same total epoch budget, the main extra work in Stage 2 is the shuffled-graph forward pass required by $\mathcal{L}_{tex}$. This cost is incurred only during training. At test time, the shuffled view is not constructed; inference uses one statistical forward pass, one graph forward pass, and logit summation.

The benefit of the schedule is visible in both convergence behavior and final metrics. Figure \ref{fig:gradient} shows that the topological gradient decays under joint training but rebounds after the statistical stream is frozen. This rebound is important: it means Stage 2 is not merely continuing the same optimization path with a different learning rate, but exposing an error signal that joint training had largely suppressed. The ablation table gives the endpoint check. Joint Opt. reaches 0.9299 AUC on PANDA and 0.9773 on TCGA-BRCA. A 10$\times$ topological learning-rate multiplier improves these values to 0.9352 and 0.9786, but the broader table shows that learning-rate balancing alone does not give the same stable, spatially sensitive solution. Curriculum scheduling and hard instance mining show the same pattern: they reduce the problem, but do not remove the competition between the streams.

The schedule also affects stability. ResTopoMIL uses a fixed residual target in Stage 2, whereas dropout and curriculum variants create a less stable target: the statistical signal is either randomly weakened or still changing while the graph branch is being trained. This does not mean two-stage training is always easier to tune. It adds one schedule boundary, requires choosing the warmup length, and depends on the statistical anchor being strong enough to remove genuine compositional signal. The main result is therefore not that two-stage training reduces engineering complexity, but that the added constraint buys a better optimization problem for topology.

\begin{table}[H]
    \centering
    \small
    \caption{\textbf{Fusion and Interaction Ablation.} Full metrics for different ways of combining statistical and topological evidence.}
    \label{tab:app_fusion_ablation}
    \setlength{\tabcolsep}{5pt}
    \renewcommand{\arraystretch}{1.06}
    \resizebox{\textwidth}{!}{
    \begin{tabular}{l ccc ccc}
        \toprule
        \multirow{2}{*}{\textbf{Variant}} & \multicolumn{3}{c}{\textbf{PANDA}} & \multicolumn{3}{c}{\textbf{TCGA-BRCA}} \\
        \cmidrule(lr){2-4}\cmidrule(lr){5-7}
        & Acc & F1 & AUC & Acc & F1 & AUC \\
        \midrule
        Gated Fusion & 0.7389$_{\pm.0068}$ & 0.6991$_{\pm.0075}$ & 0.9225$_{\pm.0062}$ & 0.9432$_{\pm.0051}$ & 0.9068$_{\pm.0065}$ & 0.9802$_{\pm.0042}$ \\
        MoE & 0.7471$_{\pm.0052}$ & 0.6918$_{\pm.0048}$ & 0.9305$_{\pm.0035}$ & 0.9432$_{\pm.0045}$ & 0.9108$_{\pm.0055}$ & \textbf{0.9897}$_{\pm.0028}$ \\
        Indep. Clf & 0.7352$_{\pm.0085}$ & 0.6845$_{\pm.0081}$ & 0.9256$_{\pm.0062}$ & 0.9465$_{\pm.0058}$ & 0.9112$_{\pm.0095}$ & 0.9845$_{\pm.0025}$ \\
        \midrule
        \textbf{ResTopoMIL} & \textbf{0.7546}$_{\pm.0094}$ & \textbf{0.7097}$_{\pm.0098}$ & \textbf{0.9426}$_{\pm.0010}$ & \textbf{0.9568}$_{\pm.0095}$ & \textbf{0.9308}$_{\pm.0166}$ & 0.9838$_{\pm.0049}$ \\
        \bottomrule
    \end{tabular}
    }
\end{table}

Table \ref{tab:app_fusion_ablation} shows that the benefit is not obtained by adding a generic interaction head. Gated Fusion performs poorly, suggesting that early adaptive mixing can reintroduce the same shortcut: the fusion gate may favor the stream that already explains the label. MoE is stronger in AUC on TCGA-BRCA, but it is weaker on PANDA AUC and TCGA-BRCA Accuracy/F1. Independent classifiers are also insufficient. The result supports the design choice that the topological stream should be trained as a residual correction, not as another branch that competes with statistics from the start.

\begin{table}[H]
    \centering
    \small
    \caption{\textbf{Architecture and Validity Ablation.} Full metrics for graph-backbone and sanity-control variants.}
    \label{tab:app_arch_validity_ablation}
    \setlength{\tabcolsep}{5pt}
    \renewcommand{\arraystretch}{1.06}
    \resizebox{\textwidth}{!}{
    \begin{tabular}{l ccc ccc}
        \toprule
        \multirow{2}{*}{\textbf{Variant}} & \multicolumn{3}{c}{\textbf{PANDA}} & \multicolumn{3}{c}{\textbf{TCGA-BRCA}} \\
        \cmidrule(lr){2-4}\cmidrule(lr){5-7}
        & Acc & F1 & AUC & Acc & F1 & AUC \\
        \midrule
        GAT (Complex) & 0.7489$_{\pm.0062}$ & 0.7074$_{\pm.0058}$ & 0.9419$_{\pm.0041}$ & 0.9432$_{\pm.0052}$ & 0.9068$_{\pm.0061}$ & \textbf{0.9889}$_{\pm.0025}$ \\
        Random Graph & 0.6815$_{\pm.0148}$ & 0.5362$_{\pm.0260}$ & 0.8286$_{\pm.0150}$ & 0.8523$_{\pm.0180}$ & 0.6749$_{\pm.0250}$ & 0.8563$_{\pm.0110}$ \\
        Fixed Proto. & 0.7335$_{\pm.0065}$ & 0.6781$_{\pm.0072}$ & 0.9349$_{\pm.0055}$ & 0.9446$_{\pm.0056}$ & 0.9098$_{\pm.0094}$ & 0.9746$_{\pm.0035}$ \\
        \midrule
        \textbf{ResTopoMIL} & \textbf{0.7546}$_{\pm.0094}$ & \textbf{0.7097}$_{\pm.0098}$ & \textbf{0.9426}$_{\pm.0010}$ & \textbf{0.9568}$_{\pm.0095}$ & \textbf{0.9308}$_{\pm.0166}$ & 0.9838$_{\pm.0049}$ \\
        \bottomrule
    \end{tabular}
    }
\end{table}

Table \ref{tab:app_arch_validity_ablation} separates architectural complexity from the proposed training logic. Replacing the 2-layer GCN with a GAT does not improve PANDA and only modestly changes TCGA-BRCA, so the gain is not explained by a more expressive graph operator. Random Graph trains the model after replacing the true neighborhood graph with random edges; its large drop in F1 and AUC shows that merely adding noisy graph connectivity is not enough. Fixed Proto. keeps the prototype assignment fixed rather than learned, and remains below ResTopoMIL on PANDA and on TCGA-BRCA Accuracy/F1. These controls support the paper's position that the contribution is not a complex graph module; it is the explicit residual formulation and the way topology is forced to remain spatial.

\begin{table}[H]
    \centering
    \small
    \caption{\textbf{Hyperparameter and Constraint Sensitivity.} Full metrics for prototype number, graph degree, margin, and texture-loss removal.}
    \label{tab:app_hparam_ablation}
    \setlength{\tabcolsep}{4.2pt}
    \renewcommand{\arraystretch}{1.04}
    \resizebox{\textwidth}{!}{
    \begin{tabular}{l ccc ccc}
        \toprule
        \multirow{2}{*}{\textbf{Variant}} & \multicolumn{3}{c}{\textbf{PANDA}} & \multicolumn{3}{c}{\textbf{TCGA-BRCA}} \\
        \cmidrule(lr){2-4}\cmidrule(lr){5-7}
        & Acc & F1 & AUC & Acc & F1 & AUC \\
        \midrule
        $K_{proto}=8$ & 0.7405$_{\pm.0071}$ & 0.6912$_{\pm.0063}$ & 0.9285$_{\pm.0051}$ & 0.9482$_{\pm.0055}$ & 0.9145$_{\pm.0089}$ & 0.9852$_{\pm.0021}$ \\
        $K_{proto}=16$ & 0.7512$_{\pm.0058}$ & 0.7065$_{\pm.0042}$ & 0.9392$_{\pm.0034}$ & 0.9515$_{\pm.0038}$ & 0.9224$_{\pm.0061}$ & 0.9889$_{\pm.0015}$ \\
        $K_{proto}=64$ & 0.7544$_{\pm.0052}$ & \textbf{0.7098}$_{\pm.0048}$ & 0.9415$_{\pm.0031}$ & 0.9528$_{\pm.0042}$ & 0.9245$_{\pm.0065}$ & 0.9892$_{\pm.0018}$ \\
        $K_{knn}=4$ & 0.7438$_{\pm.0062}$ & 0.6955$_{\pm.0059}$ & 0.9312$_{\pm.0045}$ & 0.9502$_{\pm.0049}$ & 0.9192$_{\pm.0071}$ & 0.9875$_{\pm.0019}$ \\
        $K_{knn}=16$ & 0.7495$_{\pm.0055}$ & 0.7024$_{\pm.0051}$ & 0.9368$_{\pm.0037}$ & 0.9518$_{\pm.0044}$ & 0.9218$_{\pm.0068}$ & 0.9882$_{\pm.0022}$ \\
        Margin $m=0.1$ & 0.7462$_{\pm.0060}$ & 0.6988$_{\pm.0055}$ & 0.9325$_{\pm.0048}$ & 0.9510$_{\pm.0040}$ & 0.9205$_{\pm.0062}$ & 0.9880$_{\pm.0018}$ \\
        Margin $m=0.2$ & 0.7535$_{\pm.0048}$ & 0.7082$_{\pm.0041}$ & 0.9408$_{\pm.0035}$ & 0.9532$_{\pm.0035}$ & 0.9248$_{\pm.0055}$ & 0.9896$_{\pm.0014}$ \\
        Margin $m=0.4$ & \textbf{0.7548}$_{\pm.0049}$ & 0.7094$_{\pm.0042}$ & 0.9417$_{\pm.0033}$ & 0.9536$_{\pm.0036}$ & 0.9251$_{\pm.0053}$ & \textbf{0.9898}$_{\pm.0014}$ \\
        Margin $m=0.5$ & 0.7505$_{\pm.0052}$ & 0.7042$_{\pm.0049}$ & 0.9385$_{\pm.0039}$ & 0.9525$_{\pm.0038}$ & 0.9230$_{\pm.0058}$ & 0.9890$_{\pm.0016}$ \\
        w/o $\mathcal{L}_{tex}$ & 0.7155$_{\pm.0072}$ & 0.6970$_{\pm.0081}$ & 0.9147$_{\pm.0065}$ & 0.9418$_{\pm.0054}$ & 0.9065$_{\pm.0098}$ & 0.9762$_{\pm.0038}$ \\
        \midrule
        \textbf{ResTopoMIL} & 0.7546$_{\pm.0094}$ & 0.7097$_{\pm.0098}$ & \textbf{0.9426}$_{\pm.0010}$ & \textbf{0.9568}$_{\pm.0095}$ & \textbf{0.9308}$_{\pm.0166}$ & 0.9838$_{\pm.0049}$ \\
        \bottomrule
    \end{tabular}
    }
\end{table}

Table \ref{tab:app_hparam_ablation} shows a broad but interpretable stability pattern. Very small prototype dictionaries and sparse KNN graphs underfit the statistical or spatial summaries, while larger settings approach the default and occasionally edge it in individual columns. The margin sweep is not a fragile optimum search: $m=0.2$ and $m=0.4$ are close to the default, whereas too small or too large a margin weakens residual separation. Removing $\mathcal{L}_{tex}$ is the strongest negative control among these rows, which supports the role of the shuffle constraint in keeping the residual branch aligned with spatial arrangement.

\section{Cross-Backbone Generalization on CTransPath}
\label{sec:ctranspath}

To check whether the ranking depends on the UNI encoder, we repeat the classification benchmarks with CTransPath features under the same implementation. We report full comparisons on NSCLC/TCGA-BRCA and BRACS/PANDA. ResTopoMIL remains strongest or near-strongest across the reported metrics, suggesting that the gain is not tied to a particular patch encoder.

\begin{table}[H]
    \centering
    \small
    \caption{\textbf{CTransPath Results on NSCLC and TCGA-BRCA.} Accuracy, F1, and AUC are reported as mean $\pm$ std.}
    \label{tab:ctranspath_lung_breast}
    \setlength{\tabcolsep}{4.8pt}
    \renewcommand{\arraystretch}{1.05}
    \resizebox{\textwidth}{!}{
    \begin{tabular}{l c c c c c c}
        \toprule
        \textbf{Method} & \textbf{NSCLC (Acc)} & \textbf{NSCLC (F1)} & \textbf{NSCLC (AUC)} & \textbf{TCGA-BRCA (Acc)} & \textbf{TCGA-BRCA (F1)} & \textbf{TCGA-BRCA (AUC)} \\
        \midrule
        AB-MIL &
        0.8810$_{\pm.0075}$ & 0.8510$_{\pm.0082}$ & 0.9412$_{\pm.0085}$ &
        0.9250$_{\pm.0075}$ & 0.9010$_{\pm.0115}$ & 0.9605$_{\pm.0035}$ \\
        CLAM-SB &
        0.8855$_{\pm.0092}$ & 0.8580$_{\pm.0095}$ & 0.9485$_{\pm.0075}$ &
        0.9150$_{\pm.0085}$ & 0.9085$_{\pm.0125}$ & 0.9685$_{\pm.0045}$ \\
        DS-MIL &
        0.8850$_{\pm.0115}$ & 0.8530$_{\pm.0098}$ & 0.9430$_{\pm.0092}$ &
        0.8910$_{\pm.0225}$ & 0.9110$_{\pm.0135}$ & 0.9622$_{\pm.0041}$ \\
        TransMIL &
        0.8950$_{\pm.0145}$ & 0.8655$_{\pm.0140}$ & 0.9515$_{\pm.0075}$ &
        0.9250$_{\pm.0115}$ & 0.8850$_{\pm.0175}$ & 0.9645$_{\pm.0160}$ \\
        ILRA-MIL &
        0.8820$_{\pm.0165}$ & 0.8710$_{\pm.0162}$ & 0.9485$_{\pm.0082}$ &
        0.9310$_{\pm.0065}$ & 0.8950$_{\pm.0105}$ & 0.9525$_{\pm.0115}$ \\
        MHIM-MIL &
        0.8910$_{\pm.0195}$ & 0.8550$_{\pm.0195}$ & 0.9610$_{\pm.0048}$ &
        0.9310$_{\pm.0105}$ & 0.8950$_{\pm.0165}$ & 0.9658$_{\pm.0142}$ \\
        DGR-MIL &
        0.8850$_{\pm.0235}$ & 0.8650$_{\pm.0245}$ & 0.9220$_{\pm.0250}$ &
        0.8950$_{\pm.0115}$ & 0.8950$_{\pm.0155}$ & 0.9605$_{\pm.0210}$ \\
        2DMambaMIL &
        0.8910$_{\pm.0185}$ & 0.8750$_{\pm.0215}$ & 0.9315$_{\pm.0195}$ &
        0.9350$_{\pm.0075}$ & 0.9010$_{\pm.0125}$ & 0.9640$_{\pm.0185}$ \\
        \midrule
        \textbf{ResTopoMIL (Ours)} &
        \textbf{0.9010}$_{\pm.0095}$ & \textbf{0.8955}$_{\pm.0095}$ & \textbf{0.9685}$_{\pm.0032}$ &
        \textbf{0.9420}$_{\pm.0105}$ & \textbf{0.9150}$_{\pm.0175}$ & \textbf{0.9750}$_{\pm.0052}$ \\
        \bottomrule
    \end{tabular}
    }
\end{table}

\begin{table}[H]
    \centering
    \small
    \caption{\textbf{CTransPath Results on BRACS and PANDA.} Accuracy, F1, and AUC are reported as mean $\pm$ std.}
    \label{tab:ctranspath_bracs_panda}
    \setlength{\tabcolsep}{4.8pt}
    \renewcommand{\arraystretch}{1.05}
    \resizebox{\textwidth}{!}{
    \begin{tabular}{l c c c c c c}
        \toprule
        \textbf{Method} & \textbf{BRACS (Acc)} & \textbf{BRACS (F1)} & \textbf{BRACS (AUC)} & \textbf{PANDA (Acc)} & \textbf{PANDA (F1)} & \textbf{PANDA (AUC)} \\
        \midrule
        AB-MIL &
        0.7250$_{\pm.0210}$ & 0.6415$_{\pm.0225}$ & 0.8512$_{\pm.0105}$ &
        0.6854$_{\pm.0092}$ & 0.6652$_{\pm.0085}$ & 0.9185$_{\pm.0042}$ \\
        CLAM-SB &
        0.7185$_{\pm.0195}$ & 0.6120$_{\pm.0215}$ & 0.8580$_{\pm.0125}$ &
        0.6820$_{\pm.0088}$ & 0.6610$_{\pm.0065}$ & 0.9150$_{\pm.0038}$ \\
        DS-MIL &
        0.7305$_{\pm.0285}$ & 0.6525$_{\pm.0320}$ & 0.8550$_{\pm.0152}$ &
        0.6815$_{\pm.0065}$ & 0.6612$_{\pm.0052}$ & 0.9205$_{\pm.0068}$ \\
        TransMIL &
        0.6150$_{\pm.0185}$ & 0.5255$_{\pm.0415}$ & 0.8125$_{\pm.0128}$ &
        0.6855$_{\pm.0105}$ & 0.6420$_{\pm.0150}$ & 0.9024$_{\pm.0055}$ \\
        ILRA-MIL &
        0.5900$_{\pm.0250}$ & 0.5105$_{\pm.0385}$ & 0.7850$_{\pm.0210}$ &
        0.7255$_{\pm.0080}$ & 0.6710$_{\pm.0115}$ & 0.9120$_{\pm.0045}$ \\
        MHIM-MIL &
        0.6350$_{\pm.0380}$ & 0.5510$_{\pm.0520}$ & 0.8015$_{\pm.0165}$ &
        0.6750$_{\pm.0135}$ & 0.6155$_{\pm.0185}$ & 0.8985$_{\pm.0035}$ \\
        DGR-MIL &
        0.6800$_{\pm.0350}$ & 0.6350$_{\pm.0485}$ & 0.7950$_{\pm.0305}$ &
        0.6720$_{\pm.0145}$ & 0.6185$_{\pm.0195}$ & 0.8845$_{\pm.0095}$ \\
        2DMambaMIL &
        0.6905$_{\pm.0285}$ & 0.6420$_{\pm.0465}$ & 0.8080$_{\pm.0255}$ &
        0.6255$_{\pm.0185}$ & 0.6815$_{\pm.0125}$ & 0.8912$_{\pm.0082}$ \\
        \midrule
        \textbf{ResTopoMIL (Ours)} &
        \textbf{0.7385}$_{\pm.0255}$ & \textbf{0.6952}$_{\pm.0435}$ & \textbf{0.8715}$_{\pm.0085}$ &
        \textbf{0.7350}$_{\pm.0110}$ & \textbf{0.6885}$_{\pm.0105}$ & \textbf{0.9258}$_{\pm.0025}$ \\
        \bottomrule
    \end{tabular}
    }
\end{table}

The cross-backbone result is consistent with the UNI-based main table, but it is also useful for a more specific reason. CTransPath changes the patch representation while leaving the MIL aggregator comparison unchanged. Table \ref{tab:ctranspath_lung_breast} shows that ResTopoMIL remains best on NSCLC and TCGA-BRCA across Accuracy, F1, and AUC, so the improvement is not an artifact of one particular feature extractor.

Table \ref{tab:ctranspath_bracs_panda} is even more diagnostic because BRACS and PANDA rely more heavily on tissue architecture. ResTopoMIL improves BRACS AUC from the strongest baseline value of 0.8580 to 0.8715, and improves PANDA AUC from 0.9205 to 0.9258. The BRACS F1 gain is also large, moving from 0.6525 for the strongest baseline to 0.6952. These improvements are not huge in every column, but they are consistent across accuracy, F1, and AUC, which is the expected pattern if topology provides complementary evidence rather than replacing the patch encoder.

\section{Additional Shuffle Sensitivity Across WSI Benchmarks}
\label{sec:shuffle_four_benchmarks}

Before showing the full progressive shuffling curves, we report the end-point sensitivity under complete coordinate permutation on representative WSI benchmarks. Patch embeddings are fixed, and only coordinates are permuted; a small AUC change therefore means that the model prediction is largely insensitive to the spatial graph.

\begin{table}[H]
    \centering
    \small
    \caption{\textbf{Additional Shuffle Sensitivity on WSI Benchmarks.} Patch embeddings are fixed; only coordinates are permuted. Endpoint values are shown for representative runs unless otherwise stated; tables in the main text report 5-seed mean$\pm$std.}
    \label{tab:shuffle_multi_dataset}
    \setlength{\tabcolsep}{5pt}
    \renewcommand{\arraystretch}{1.05}
    \resizebox{\textwidth}{!}{
    \begin{tabular}{lccc}
        \toprule
        \textbf{Method} & \textbf{BRACS} & \textbf{PANDA} & \textbf{NSCLC} \\
        & Orig $\rightarrow$ Shuff ($\Delta$) & Orig $\rightarrow$ Shuff ($\Delta$) & Orig $\rightarrow$ Shuff ($\Delta$) \\
        \midrule
        AB-MIL & 0.8806$\rightarrow$0.8810 (+0.0004) & 0.9306$\rightarrow$0.9302 (-0.0004) & 0.9569$\rightarrow$0.9571 (+0.0002) \\
        TransMIL & 0.8450$\rightarrow$0.8425 (-0.0025) & 0.9288$\rightarrow$0.9270 (-0.0018) & 0.9692$\rightarrow$0.9675 (-0.0017) \\
        DS-MIL & 0.8054$\rightarrow$0.8054 (0.0000) & 0.9329$\rightarrow$0.9329 (0.0000) & 0.9579$\rightarrow$0.9579 (0.0000) \\
        ResTopoMIL & 0.9006$\rightarrow$0.8350 (-0.0656) & 0.9426$\rightarrow$0.9052 (-0.0374) & 0.9753$\rightarrow$0.9610 (-0.0143) \\
        \bottomrule
    \end{tabular}
    }
\end{table}

Table \ref{tab:shuffle_multi_dataset} gives a compact view of spatial sensitivity at the 100\% shuffle endpoint. AB-MIL is almost unchanged, as expected for a permutation-invariant model. DS-MIL is exactly unchanged on the listed benchmarks, and TransMIL shows only mild degradation despite modeling contextual relations. ResTopoMIL drops more clearly when coordinates are destroyed, especially on BRACS and PANDA. Together with the progressive curves below, this table separates slide-level predictive strength from actual coordinate dependence: high AUC alone does not imply that the decision rule is using tissue arrangement.

\section{Progressive Coordinate-Shuffling Analysis}
\label{sec:shuffle_appendix}

\begin{figure}[H]
    \centering
    \includegraphics[width=\textwidth]{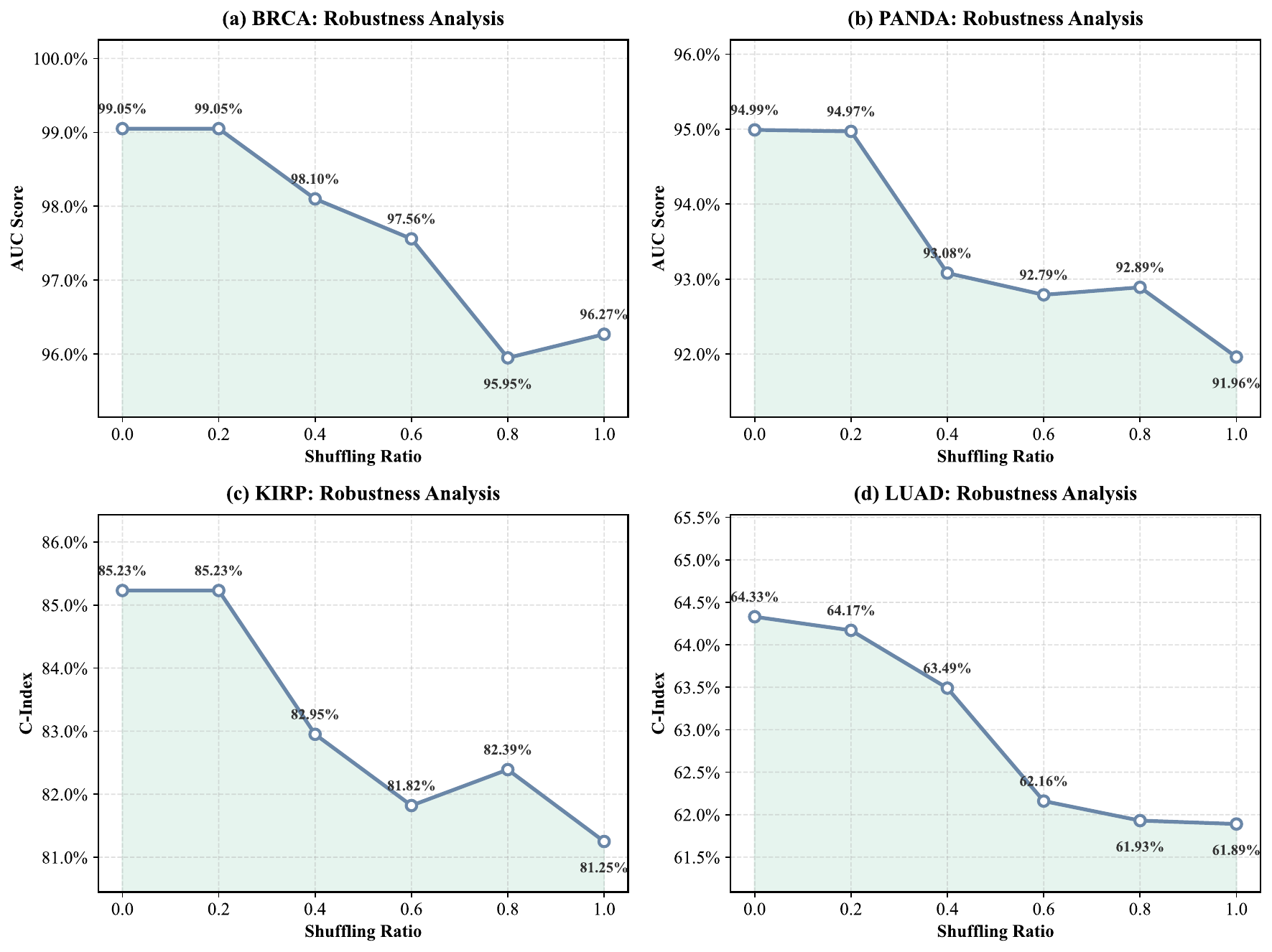}
    \caption{\textbf{Progressive Shuffle Sensitivity.} AUC decreases as an increasing fraction of patch coordinates is permuted, while the patch composition remains unchanged. Curves show representative-seed trajectories for trend visualization; numeric tables report 5-seed mean$\pm$std unless explicitly marked otherwise.}
    \label{fig:shuffling_app}
\end{figure}

Figure \ref{fig:shuffling_app} reports the full progressive coordinate-shuffling analysis. This experiment keeps patch embeddings fixed and gradually corrupts only the spatial coordinates used to construct context. The resulting monotonic degradation provides a behavioral check that complements the ablations in the main text: the residual branch depends on preserved spatial arrangement rather than only on extra capacity or a favorable optimization schedule.

Figure \ref{fig:shuffling_app} adds a dose-response view to the endpoint table. The important property is not only that the fully shuffled model performs worse, but that the degradation increases as a larger fraction of coordinates is permuted. This monotonic trend is a stronger check than a single shuffle point because it shows that the model response tracks the amount of spatial damage. In contrast, a model that uses coordinates only incidentally would not be expected to produce a smooth decline as the coordinate field is progressively corrupted.

The shuffling test should not be read as evidence that the model is fragile to ordinary coordinate noise. ResTopoMIL operates on the coordinate-induced KNN graph, not on absolute slide coordinates. Rigid translation, rotation, reflection, and uniform scaling preserve pairwise distances and therefore leave the unweighted KNN graph unchanged. Mild coordinate jitter is also unlikely to affect the prediction when the local neighbor identities remain stable. The expected failure mode is different: performance should drop when the perturbation is large enough to rewire local neighborhoods and erase the tissue arrangement that supports the diagnosis. Figure \ref{fig:shuffling_app} is consistent with this selective behavior. Small topology-preserving changes should be tolerated, whereas progressive coordinate permutation degrades performance because it increasingly destroys the graph structure. A dedicated invariance study over registration noise and rigid transformations remains useful future work.

\section{Quantitative Localization Assessment on CAMELYON-16}
\label{sec:camelyon_localization}
\label{sec:camelyon_protocol}

To further examine whether overcoming spatial blindness translates into more useful spatial evidence, patch-level lesion localization is evaluated on CAMELYON-16 \citep{ehteshami2017diagnostic}. The official tumor annotations are used only for evaluation; training remains slide-level, and tumor masks are not used to supervise any MIL model. Because patch-level localization scores are sensitive to mask conversion and thresholding, we specify the evaluation protocol before reporting the numbers.

CAMELYON-16 is used here as an additional localization benchmark and is not counted among the 9 primary WSI benchmarks used for classification and survival prediction.

\paragraph{Patch-level labels.}
Slides are tiled with the same non-overlapping $256\times256$ patches at 20$\times$ magnification used elsewhere in the paper. Each patch footprint is mapped to the coordinate system of the CAMELYON-16 annotation mask. A patch is labeled positive if at least 1\% of its area overlaps an annotated tumor region, and negative otherwise. This small-overlap rule includes boundary patches while avoiding isolated mask-contact artifacts. Patches outside the tissue mask produced by Otsu thresholding are ignored for both prediction and evaluation. The official annotations are therefore converted into a binary patch grid, rather than into pixel-level supervision.

\paragraph{Patch-level scores.}
For ResTopoMIL, patch evidence is extracted from the node-level topological residual before graph pooling. Let $\mathbf{H}^{(2)}_i$ be the final GCN embedding of patch $i$ and let $\mathbf{W}_{topo,y}$ be the residual-classifier weight for the target or positive class. We use
\begin{equation}
    s_i=(\mathbf{W}_{topo,y})^\top \mathbf{H}^{(2)}_i
\end{equation}
as the raw patch-level localization score, then min--max normalize scores within each slide for heatmap rendering and threshold-based evaluation. Baseline heatmaps use their native attention or instance-score outputs under the same normalization and evaluation protocol.

\paragraph{Dice threshold.}
Dice is computed on the patch grid. The binarization threshold is selected on the validation split by maximizing mean Dice for each method, and the selected threshold is then fixed for test slides. We do not tune the threshold on test slides or choose slide-specific thresholds.

\paragraph{FROC computation.}
FROC is threshold-swept rather than evaluated at a single operating point. For each threshold, connected components in the binarized patch grid are treated as lesion candidates, and the candidate score is the maximum patch score inside the component. A candidate is counted as a true positive if its maximum-score patch falls inside an annotated tumor region; otherwise it is counted as a false positive. The reported FROC is the average sensitivity at the standard CAMELYON operating points of $1/8$, $1/4$, $1/2$, $1$, $2$, $4$, and $8$ false positives per slide. All methods use the same patch labels, threshold-selection protocol, component rule, and FROC operating points.

With this fixed protocol, Table \ref{tab:camelyon_loc} reports Dice, specificity, and FROC. TransMIL illustrates a mismatch between false-positive control and lesion overlap: it achieves near-perfect specificity (0.999), but its localization performance is weak by Dice (0.103) despite a moderate FROC (0.4866). This is consistent with the optimization-laziness interpretation: a model can suppress false positives while still failing to assign spatial evidence to the correct tumor boundary.

\begin{table}[H]
    \centering
    \small
    \caption{\textbf{Quantitative Localization on CAMELYON-16.} Patch-level Dice, specificity, and FROC are reported. Higher is better for all metrics.}
    \label{tab:camelyon_loc}
    \begin{tabular}{lccc}
        \toprule
        \textbf{Method} & \textbf{Dice} & \textbf{Specificity} & \textbf{FROC} \\
        \midrule
        AB-MIL & 0.412 & 0.985 & 0.3952 \\
        CLAM-SB & 0.459 & 0.987 & 0.4257 \\
        DS-MIL & 0.259 & 0.863 & 0.4506 \\
        TransMIL & 0.103 & \textbf{0.999} & 0.4866 \\
        DTFD-MIL & 0.525 & \textbf{0.999} & 0.4712 \\
        MHIM-MIL & 0.548 & 0.992 & 0.4815 \\
        2DMambaMIL & 0.475 & 0.995 & 0.4520 \\
        \textbf{ResTopoMIL} & \textbf{0.624} & \textbf{0.999} & \textbf{0.5483} \\
        \bottomrule
    \end{tabular}
\end{table}

ResTopoMIL achieves the best localization performance while maintaining near-perfect specificity. Its Dice score reaches 0.624 and FROC reaches 0.5483. By offloading compositional statistics to the statistical stream, the topological branch acts as a spatial regularizer and is encouraged to respect local adjacency in the tumor microenvironment. This result also clarifies why specificity alone is insufficient for evaluating structure-aware pathology models: avoiding false-positive activations can coexist with poor spatial overlap. The main conclusion from Table \ref{tab:camelyon_loc} should therefore be read as evidence that ResTopoMIL produces more spatially coherent weakly supervised evidence under a fixed protocol, not as a claim that it is a fully validated tumor segmentation system.

\section{Feature-Space Visualization of Statistical and Topological Streams}
\label{sec:feature_space_vis}

The feature spaces learned by the statistical and topological streams on BRACS are visualized using both principal component analysis (PCA) and t-SNE. These two projections provide complementary views. PCA is a linear projection that preserves the dominant global variance directions, and is therefore useful for checking whether the separation between streams is visible without relying on a nonlinear embedding method. t-SNE, by contrast, emphasizes local neighborhood structure and is better suited for inspecting whether samples from the same class form compact local clusters and whether difficult samples are reorganized by the topological stream.

\begin{figure}[H]
    \centering
    \includegraphics[width=\textwidth]{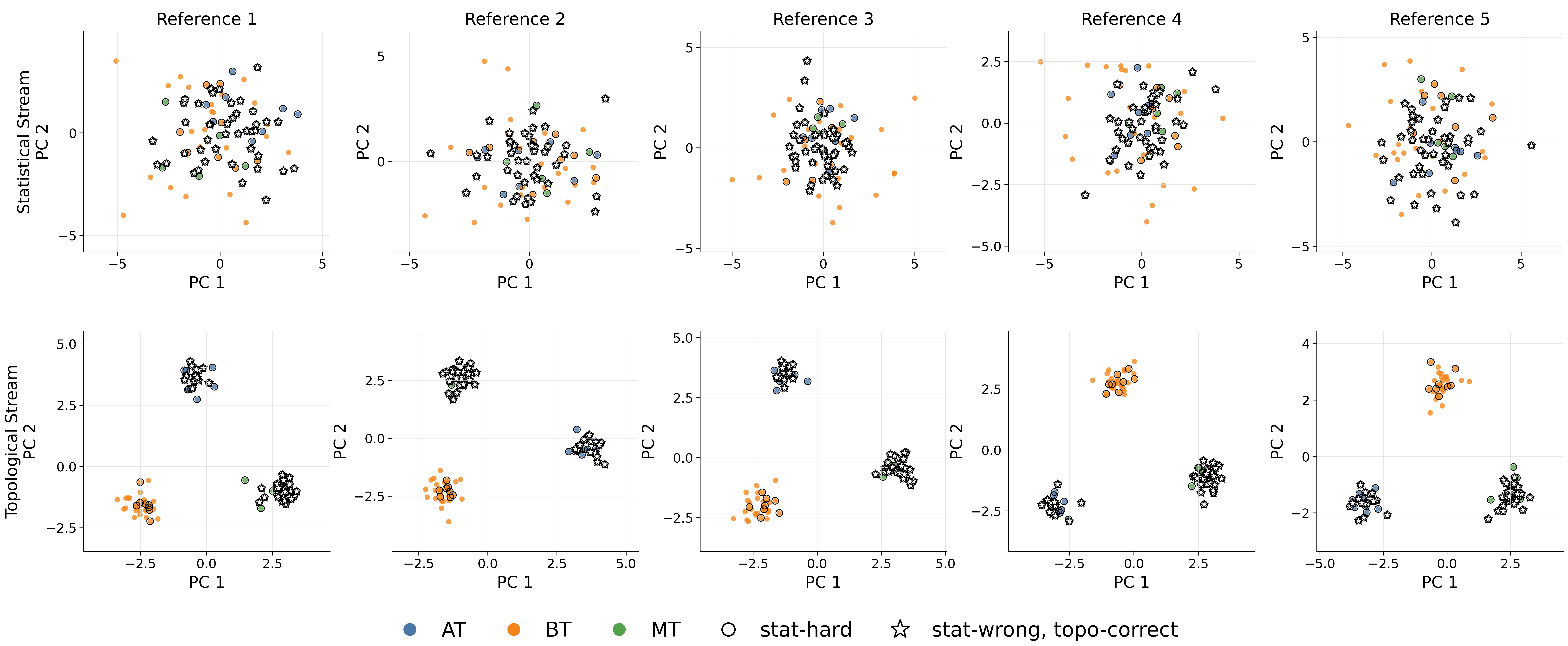}
    \caption{\textbf{PCA Visualization of Statistical and Topological Streams.} The statistical stream produces heavily mixed class distributions, with negative silhouette scores in the shown projections (e.g., around $-0.05$ to $-0.09$) and large Davies--Bouldin (DB) indices. In contrast, the topological stream forms much clearer class-separated structures, with silhouette scores around $0.87$--$0.89$ and DB indices around $0.16$--$0.18$. Hollow circles denote statistically hard samples, and stars denote samples misclassified by the statistical stream but corrected by the topological stream.}
    \label{fig:pca_stream_vis}
\end{figure}

Although PCA is only a two-dimensional linear projection, Figure \ref{fig:pca_stream_vis} already separates the two streams clearly. The statistical stream contains large overlaps among BRACS classes, matching the intuition that patch-composition summaries struggle with atypical and borderline lesions. The topological stream, by contrast, forms separated class regions with much better silhouette and DB indices. The corrected samples marked by stars are especially important: they show that the topological stream is not merely producing a prettier embedding, but reorganizing cases that are difficult for the statistical stream.

\begin{figure}[H]
    \centering
    \includegraphics[width=\textwidth]{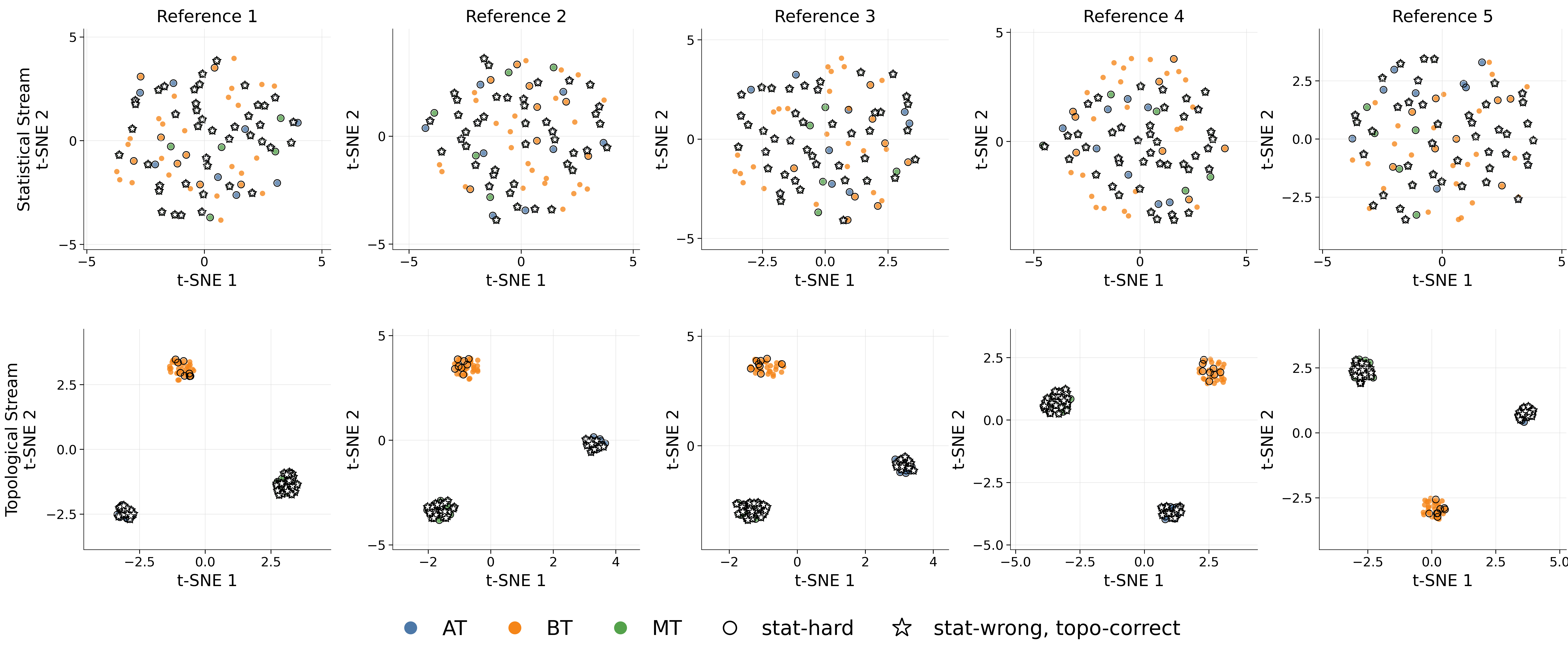}
    \caption{\textbf{t-SNE Visualization of Statistical and Topological Streams.} The nonlinear projection shows the same qualitative pattern as PCA. Statistical-stream features remain entangled, with negative silhouette scores and large DB indices, whereas topological-stream features form compact class-specific clusters with silhouette scores around $0.92$--$0.93$ and DB indices close to $0.10$. Many star-marked cases move from mixed regions in the statistical stream to the corresponding class clusters in the topological stream.}
    \label{fig:tsne_stream_vis}
\end{figure}

Figure \ref{fig:tsne_stream_vis} confirms the same separation from a local-neighborhood perspective. The t-SNE projection is useful here because it emphasizes whether samples with similar learned representations form compact neighborhoods. The statistical stream again leaves many classes entangled, while the topological stream places corrected samples near their true-class clusters. Since the PCA figure shows a compatible trend, the conclusion does not rely only on nonlinear t-SNE visualization.

Together, the two visualizations indicate that the statistical stream and the topological stream encode different information. In the statistical stream, AT, BT, and MT samples are substantially intermixed, and the negative silhouette scores together with high DB values indicate poor class separation. This is expected because the statistical stream mainly captures compositional and texture-distribution information, which can be insufficient for fine-grained BRACS categories whose distinction depends on subtle tissue organization.

The topological stream shows a markedly different structure. Across both PCA and t-SNE, the topological features yield compact and well-separated clusters, and the quantitative clustering indicators improve sharply. Importantly, the star-marked samples, which correspond to cases misclassified by the statistical stream but corrected by the topological stream, tend to move from ambiguous mixed regions into the neighborhoods of their true classes. The hollow markers also highlight statistically hard samples that are better organized after incorporating spatial topology.

These observations support the intended decoupling behavior of ResTopoMIL. PCA shows that the separation is visible even from a global linear perspective, while t-SNE confirms that the local neighborhood structure is also improved. Therefore, the apparent class separation is not merely an artifact of a nonlinear visualization method; rather, both projections suggest that the topological stream learns complementary spatial-organization cues that help resolve samples that are ambiguous under purely statistical representations.

\section{Qualitative Analysis}
\label{sec:vis}

Attention heatmaps are used as qualitative, pathology-reviewed evidence rather than pixel-level validation, since the model is trained with slide-level labels. The visual findings in this section were reviewed and confirmed by pathologists as qualitatively consistent with tumor-relevant regions and diagnostically meaningful tissue organization.

\paragraph{Heatmap scores and post-processing.}
For each method, the attention or localization score assigned to a retained patch is linearly normalized to $[0,1]$ within each slide before visualization. We do not apply CRF refinement, Gaussian smoothing, morphological closing, or tumor-shape priors. The only post-processing step is the removal of non-tissue patches using the same tissue mask as preprocessing. This choice is intentionally conservative: the displayed heatmaps should reflect the MIL scoring function rather than a separate segmentation pipeline.

\begin{figure}[H]
    \centering
    \begin{subfigure}[t]{0.49\textwidth}
        \centering
        \includegraphics[width=\linewidth]{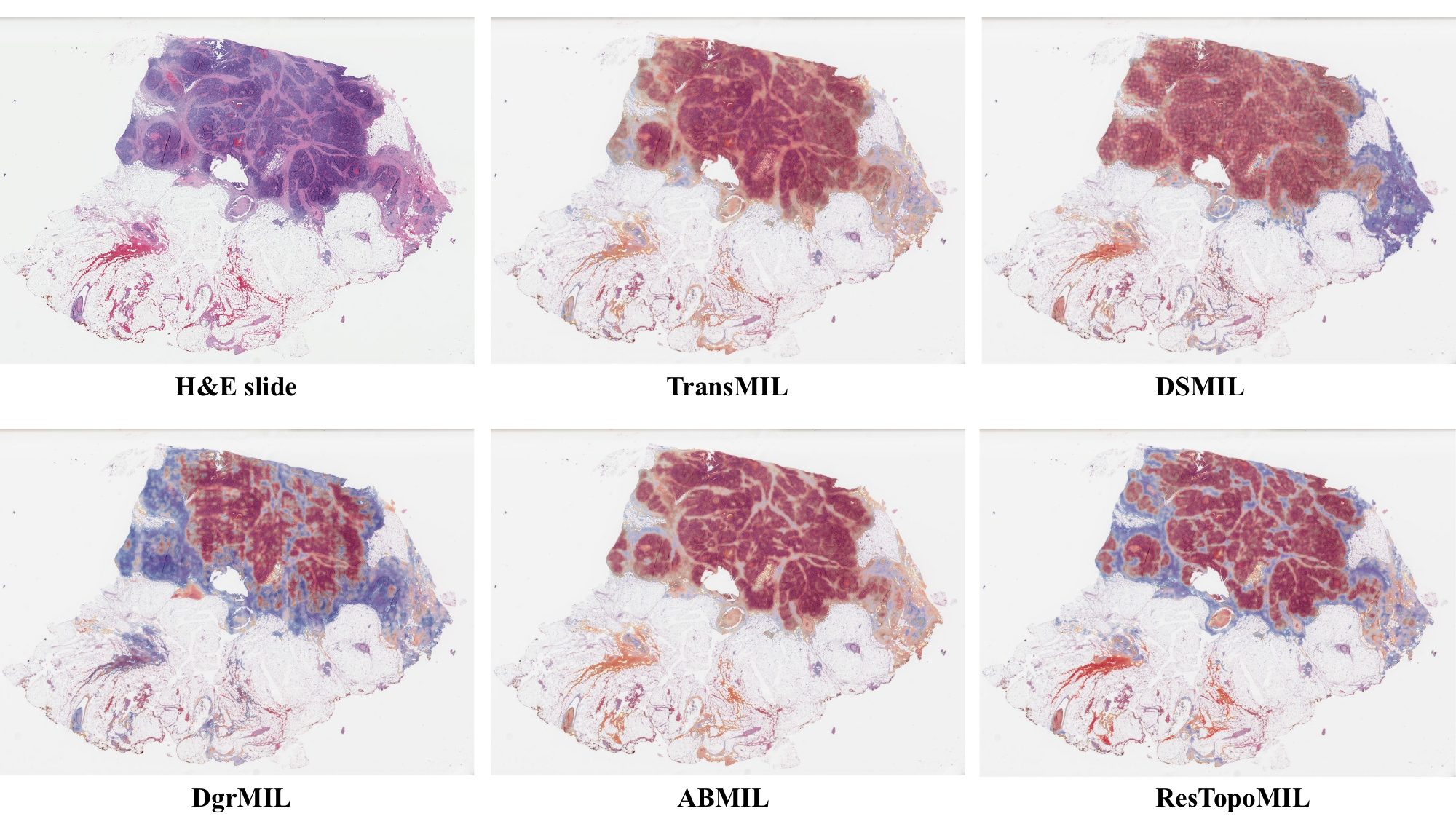}
        \caption{Representative case}
        \label{fig:attnmap_app_main}
    \end{subfigure}
    \hfill
    \begin{subfigure}[t]{0.49\textwidth}
        \centering
        \includegraphics[width=\linewidth]{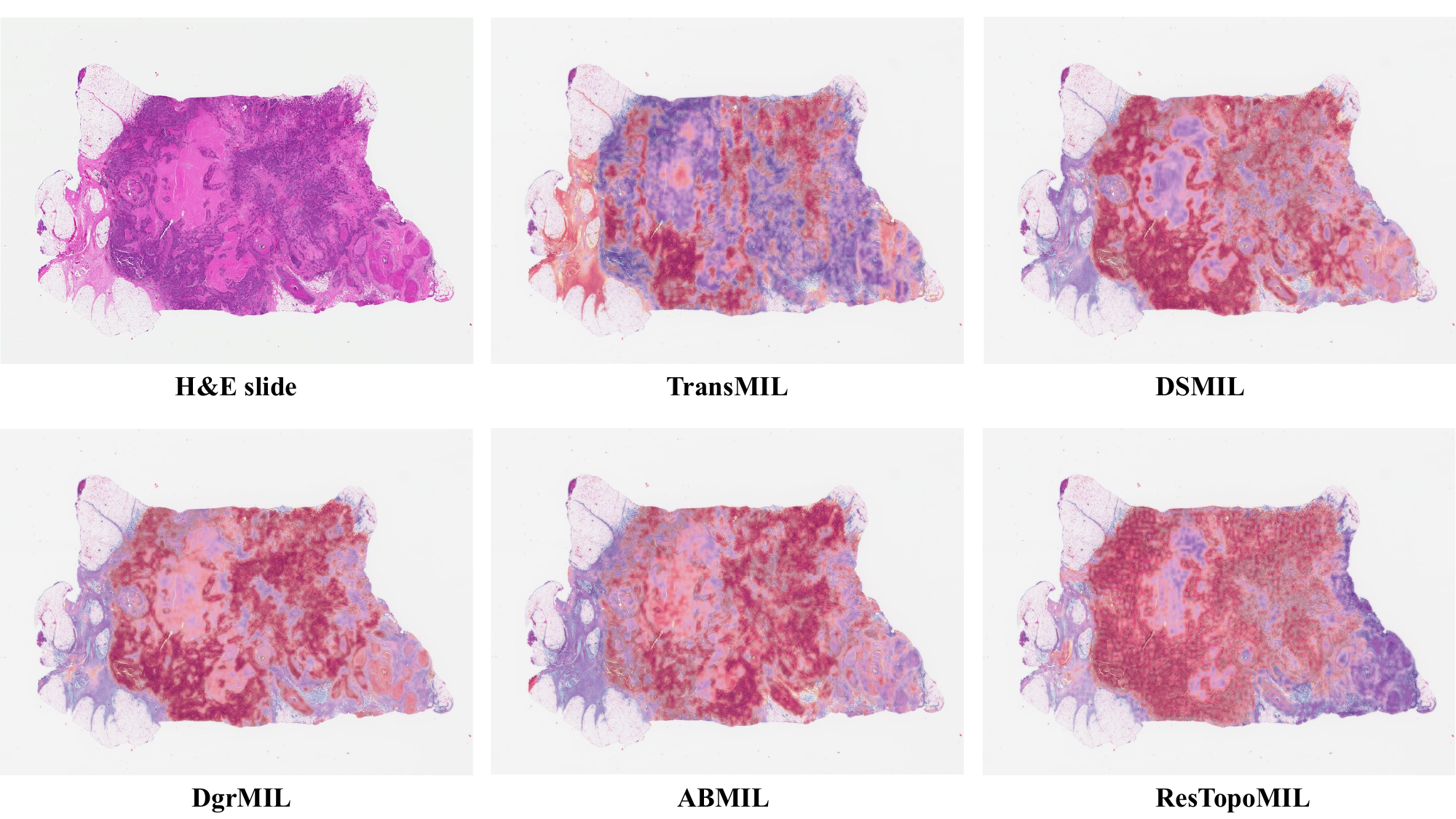}
        \caption{TCGA-A2-A3XY}
        \label{fig:sample1}
    \end{subfigure}
    \vspace{2pt}
    \begin{subfigure}[t]{0.49\textwidth}
        \centering
        \includegraphics[width=\linewidth]{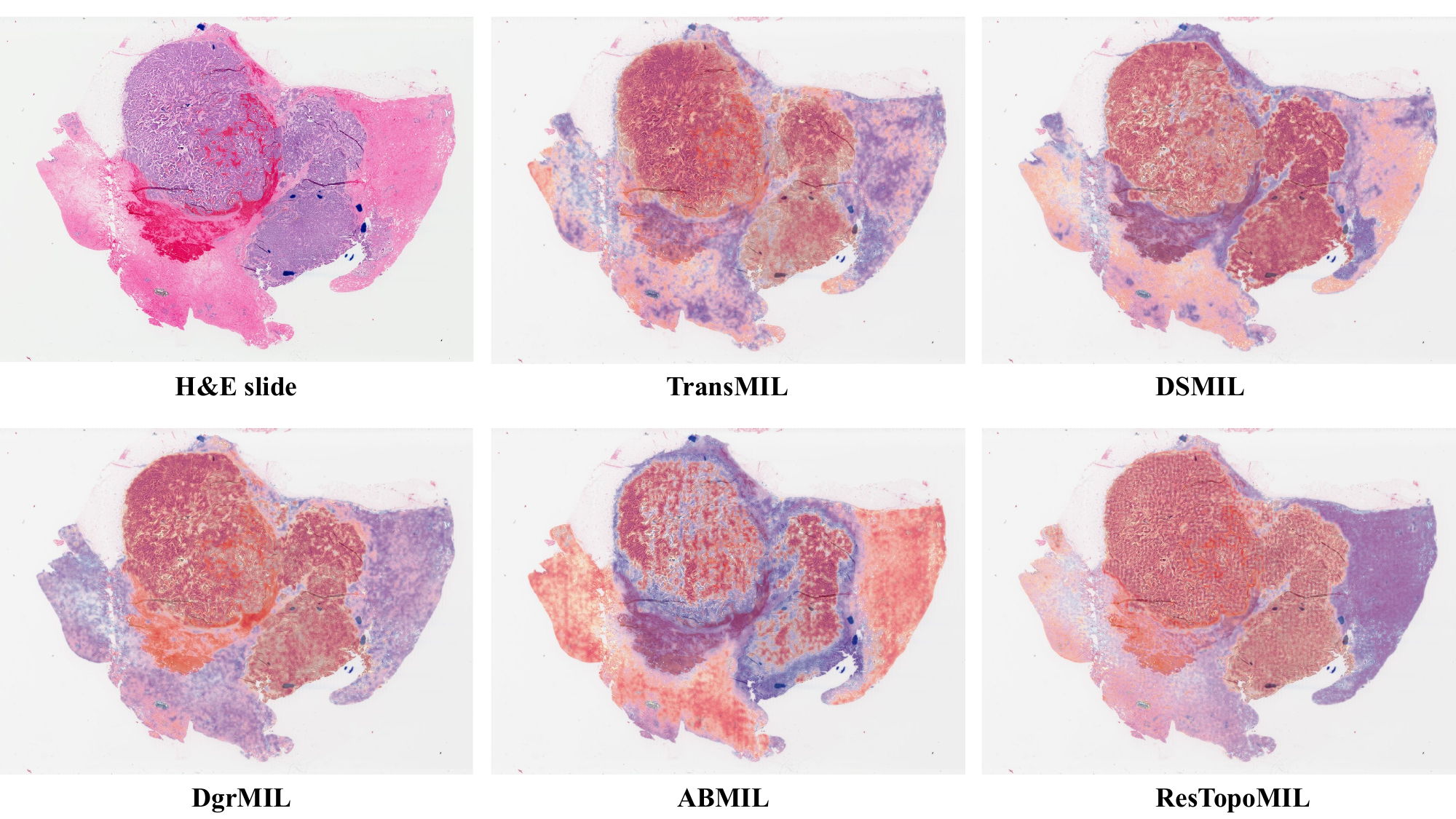}
        \caption{TCGA-B6-A0IA}
        \label{fig:sample2}
    \end{subfigure}
    \hfill
    \begin{subfigure}[t]{0.49\textwidth}
        \centering
        \includegraphics[width=\linewidth]{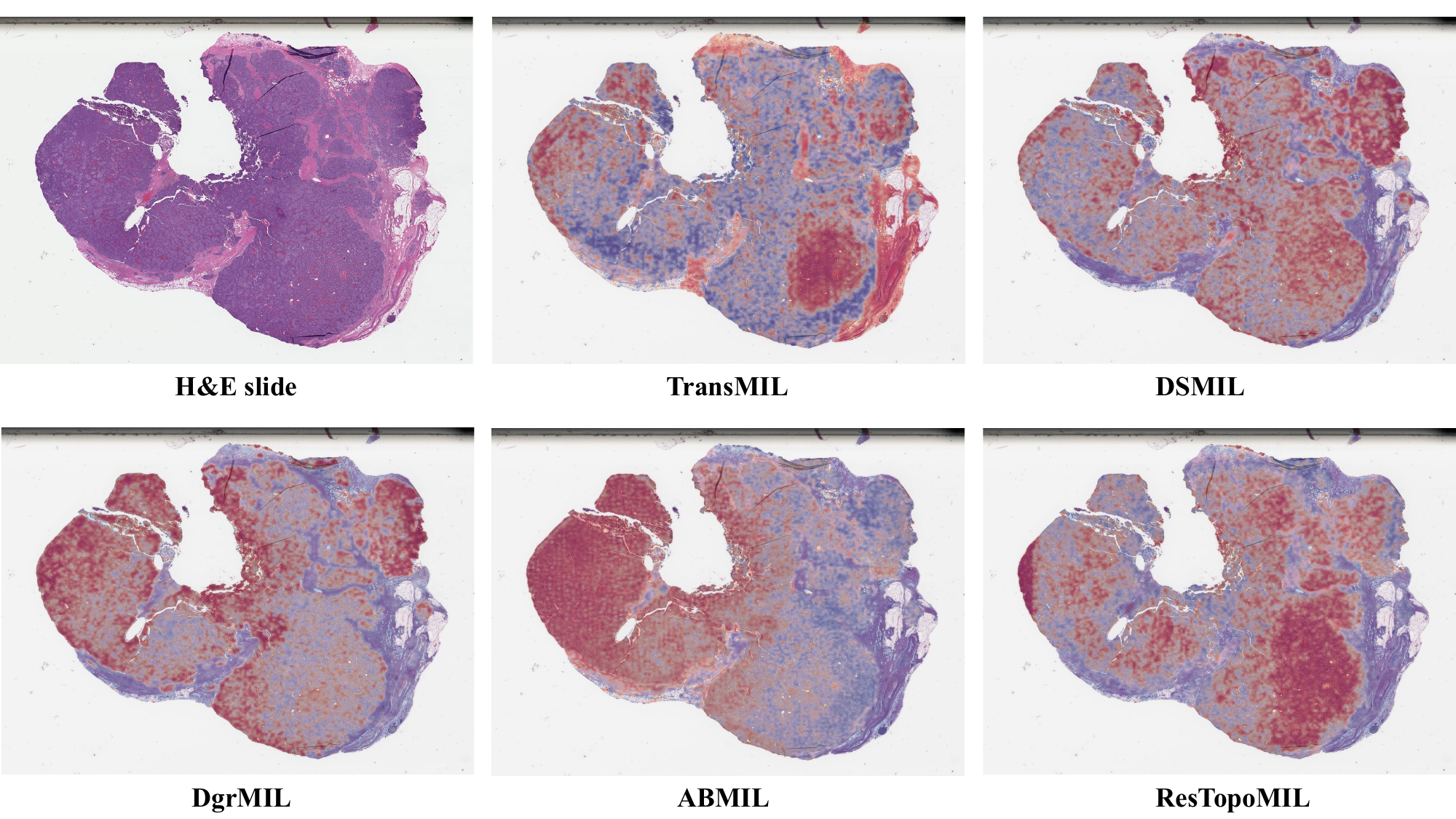}
        \caption{TCGA-LL-A442}
        \label{fig:sample3}
    \end{subfigure}
    \caption{\textbf{Attention Heatmap Visualization.} Representative and additional pathology-reviewed heatmaps. Warmer colors indicate higher attention weights. Compared with TransMIL, ResTopoMIL shows less background leakage and more contiguous attention over tumor-relevant regions.}
    \label{fig:attnmap_app}
\end{figure}

Figure \ref{fig:attnmap_app} collects the representative comparison discussed in the main text and three additional TCGA examples. The representative case (Fig. \ref{fig:attnmap_app_main}) shows more contiguous tumor attention with less diffuse background activation. \textbf{TCGA-A2-A3XY} (Fig. \ref{fig:sample1}) constrains high attention to the tumor core, avoiding stromal leakage. \textbf{TCGA-B6-A0IA} (Fig. \ref{fig:sample2}) gives sharper boundary delineation than TransMIL. \textbf{TCGA-LL-A442} (Fig. \ref{fig:sample3}) localizes invasive-front regions more consistently, suggesting that the topological branch captures architectural arrangement beyond local texture.

\section{Additional Limitations and Negative-Result Scope}
\label{sec:additional_limitations}

The experiments deliberately emphasize structure-dependent WSI tasks, because those are the settings where spatial blindness is most clinically and methodologically relevant. This focus leaves a weaker view of the opposite regime: real pathology tasks that are almost purely compositional. In such tasks, a strong statistical stream should be close to optimal, and a residual topological branch could plausibly add noise or overfit incidental tissue layout.

Spatial-MNIST-Bag Dataset A provides a controlled synthetic check of this regime, showing that ResTopoMIL can still solve a pure-composition MIL problem. However, it is not a realistic WSI benchmark. We currently lack a public real WSI dataset whose label is known to be determined primarily by composition while being insensitive to tissue arrangement. Because of this missing negative-control dataset, our exploration of composition-dominant failure modes remains limited. A useful future benchmark would contain real WSI labels for which pathologists agree that topology is largely irrelevant; such a dataset would test whether ResTopoMIL correctly defaults to statistical performance without introducing unnecessary residual variance.

The two-stage training protocol is another limitation. It makes the optimization target cleaner for the graph branch, but it also introduces a schedule choice: the statistical anchor must be trained long enough to capture composition, yet not treated as a perfect model. If the anchor is weak, Stage 2 may be asked to correct errors that are not truly spatial. If the anchor overfits, the residual available to the graph branch may be noisy or too small. The present experiments use a fixed 10-epoch warmup and then 30/20 epochs of refinement, and the ablations show that this choice is stable on the evaluated benchmarks. Still, the method is less plug-and-play than a single end-to-end MIL baseline, and future work should study adaptive stopping rules or validation criteria for deciding when to freeze the statistical stream.

\end{document}